\documentclass[10pt,twocolumn,letterpaper]{article}

\usepackage{iccv} %
\usepackage[table,x11names]{xcolor}
\usepackage[utf8]{inputenc} %
\usepackage[T1]{fontenc}    %

\usepackage{graphicx}
\usepackage{tikz}
\usetikzlibrary{positioning,shapes,arrows,fit,decorations.pathreplacing,patterns,calc,backgrounds}
\usepackage{multirow}
\usepackage{pgfplots} 
\usepackage{pgfplotstable}
\usepgfplotslibrary{polar}

\pgfplotsset{compat=1.18}

\usepackage{rotating}
\usepackage{verbatim}
\usepackage{relsize}
\usepackage{booktabs}
\usepackage[normalem]{ulem}

\usepackage{microtype}
\usepackage{etoolbox}

\definecolor{cvprblue}{rgb}{0.21,0.49,0.74}
\usepackage[pagebackref,breaklinks,colorlinks,allcolors=cvprblue]{hyperref}

\def\confYear{2025}

\newcommand\cls{\texttt{[cls]}}

\title{
        Beyond \cls{}: Exploring the true potential\\of Masked Image Modeling representations
}

\author{Marcin Przewięźlikowski$^{1,2,*}$ \quad Randall Balestriero$^{3}$ \quad  Wojciech Jasiński$^{1,4}$\\
Marek Śmieja$^{1}$\quad  Bartosz Zieliński$^{1}$\\
$^{1}$Faculty of Mathematics and Computer Science, Jagiellonian University \\
$^{2}$Doctoral School of Exact and Natural Sciences, Jagiellonian University\\
$^{3}$Brown University \quad $^{4}$AGH University of Krakow \\
 $^*${\tt\small marcin.przewiezlikowski@doctoral.uj.edu.pl}
}

\pgfplotsset{compat=1.18}

\usepackage{pgfplots} 

\usepackage{pgfplots}
\pgfplotsset{compat=newest}

\definecolor{cyan}{RGB}{0,255,255}
\definecolor{darkslategray38}{RGB}{38,38,38}
\definecolor{green}{RGB}{0,128,0}
\definecolor{lavender234234242}{RGB}{244,244,252}

\definecolor{lightgray204}{RGB}{204,204,204}
\definecolor{orange}{RGB}{255,165,0}
\definecolor{purple}{RGB}{128,0,128}

\pgfplotsset{
    sec4style/.style={
        axis background/.style={fill=lavender234234242},
        width=\linewidth,
        height=0.7\linewidth,
        axis line style={white},
        legend cell align={left},
        legend columns=3,
        legend style={
          fill opacity=0.8,
          draw opacity=1,
          text opacity=1,
          font=\footnotesize,
          at={(1,-0.4)},
          anchor=east,
          draw=lightgray204,
          fill=lavender234234242,
          /tikz/column 2/.style={anchor=base, text width=0.25\linewidth} %
        },
        tick align=outside,
        x grid style={dotted},
        y grid style={dotted},
        xmajorgrids,
        xmajorticks=true,
        xtick style={color=black},
        xtick pos=left, %
        ymajorgrids,
        ymajorticks=true,
        ytick style={color=black},
        ytick pos=left,
        scaled y ticks=false, %
        y tick label style={/pgf/number format/fixed, /pgf/number format/precision=2}, %
        tick label style={font=\footnotesize, color=black},
        legend entries={
            {DINO~\cite{caron2021emerging}},
            {iBOT~\cite{zhou2022image}},
            {MAE~\cite{he2021masked}},
            {MoCo-v3~\cite{chen2021empirical}},
            {Supervised~\cite{dosovitskiy2021image}},
            {MAE (FT)~\cite{he2021masked}},
        },
    }
}

\pgfplotsset{
    MoCoStyle/.style={
        very thick, Cyan3, mark=none, %
    },
    SupervisedStyle/.style={
        very thick, DarkOrange2, mark=none, dashed
    },
    DINOStyle/.style={
        very thick, purple, mark=none, %
    },
    iBOTStyle/.style={
        very thick, blue, mark=none, %
    },
    MAEStyle/.style={
        ultra thick, green,  mark=*, mark size=1pt %
    },
    FinetunedMAEStyle/.style={
        ultra thick, red,  mark=diamond*, mark size=1pt, %
    }
}

\begin{document}

\maketitle

\begin{abstract}
Masked Image Modeling (MIM) has emerged as a promising approach for Self-Supervised Learning (SSL) of visual representations. However, the out-of-the-box performance of MIMs is typically inferior to competing approaches. Most users cannot afford fine-tuning due to the need for large amounts of data, high GPU consumption, and specialized user knowledge. Therefore, the practical use of MIM representations is limited.
In this paper we ask what is the reason for the poor out-of-the-box performance of MIMs. Is it due to weaker features produced by MIM models, or is it due to suboptimal usage?
Through detailed analysis, we show that attention in MIMs is spread 
almost uniformly 
over many patches, leading to ineffective aggregation by the \cls{} token. 
Based on this insight, we propose Selective Aggregation to better capture the rich semantic information retained in patch tokens, which significantly improves the out-of-the-box performance of MIM\footnote{We release the codebase at 
\href{https://github.com/gmum/beyond_cls}{\url{github.com/gmum/beyond_cls}}.}.
\end{abstract}

\section{Introduction}

\begin{figure}[t]
    \centering
    \resizebox{\linewidth}{!}{
    \begin{tikzpicture}

\begin{axis} [
    axis background/.style={fill=lavender234234242},
    ybar,
    bar width=15pt,
    width=\linewidth,
    height=0.8\linewidth,
    ymin=63,
    ymax=92,
    axis y line*=left,
    axis x line*=bottom,
    ymajorgrids=true,
    grid style=dotted,
    axis line style={white},
    symbolic x coords={
        MAE~\cite{he2021masked},
        I-JEPA~\cite{assran2023ijepa},
        BEiT-v2~\cite{peng2022beitv2maskedimage},
        CAPI~\cite{darcet2025clusterpredictlatentspatches},
    }, 
    x tick label style={rotate=1},
    xtick=data,
    ylabel={ImageNet-1k accuracy},
    x=0.25\linewidth,
    enlarge x limits={0.1, 0.2},
    legend entries={
        Standard representation,
        Selective token aggregation
    },
    legend image code/.code={%
      \draw[#1] (0.cm,-0.25cm) rectangle (0.6cm,0.15cm);
    },
    legend style={
        inner ysep=2mm,              %
    },
    clip=false
    ]

\addplot[fill=red, postaction={pattern=horizontal lines}] coordinates {
(MAE~\cite{he2021masked},67.8) 
(I-JEPA~\cite{assran2023ijepa}, 77.7)
(BEiT-v2~\cite{peng2022beitv2maskedimage},78.9)
(CAPI~\cite{darcet2025clusterpredictlatentspatches}, 76.2)
}; 

\addplot[fill=Green3] coordinates {
(MAE~\cite{he2021masked}, 71.6) 
(I-JEPA~\cite{assran2023ijepa}, 79.2)
(BEiT-v2~\cite{peng2022beitv2maskedimage}, 81.0)
(CAPI~\cite{darcet2025clusterpredictlatentspatches}, 82.4)
};

\end{axis}

\end{tikzpicture}
    }
    \caption{
    The standard approaches used to obtain global representations in Masked Image Modeling (MIM) -- \cls{} token or naive averaging over patch tokens -- do not focus on the most relevant image fragments, resulting in poor out-of-the-box performance. As a remedy, we propose \textbf{Selective Aggregation} -- a lightweight approach that dynamically selects relevant tokens, thereby improving performance.
    }
    \label{fig:teaser_barplot}
    \vspace{-1\baselineskip}
\end{figure}

\newcommand{\GreenTrapezium}{\tikz[baseline]{\node[trapezium, trapezium stretches, trapezium angle=105, draw, fill=green, minimum width=0.36cm, minimum height=0.66\baselineskip, yshift=0.2\baselineskip] {};}}

\begin{figure*}[t]
    \centering
    \resizebox{\linewidth}{!}{
            
            \input{img/teaser/exec_teaser_marek.tikz}
            }
    \caption{
    ViTs trained with {\color{blue!50}Joint-Embedding Architectures (JEA)} attend to semantically rich patches while forming global \cls{} representations, which is critical for perception performance. At the same time, ViTs trained with {\color{red!50}Masked Image Modeling (MIM)} attend more uniformly to all patches, absorbing both relevant and irrelevant information and achieving an effect similar to naive average pooling (see \textbf{left} and \textbf{center}). To improve out-of-the-box MIM performance, we propose {\color{green}Selective Aggregation} (see \textbf{right}) -- a mechanism that aggregates patch tokens according to their relevance, as quantified by a lightweight linear regressor 
    (\protect\GreenTrapezium).
    }  
    \label{fig:large_teaser}
    \vspace{-1\baselineskip}
\end{figure*}

Self-supervised Learning (SSL)~\cite{balestriero2023cookbook} has emerged as a powerful paradigm for pre-training visual representations from unlabelled data. These representations are of high quality and can be used out-of-the-box for various downstream tasks~\cite{he2021masked,caron2021emerging,oquab2023dinov2,assran2023ijepa}, which is crucial because the computational costs and data volumes required for fine-tuning are prohibitive for most end users~\cite{oquab2023dinov2}. However, to take full advantage of these representations, we need to understand their distinct properties.

There are two dominant SSL paradigms: Joint Embedding Architectures (JEA), which optimize the goal of producing similar embeddings from multiple views of the same image~\cite{he2020momentum,chen2020simple,chen2020improved,caron2020unsupervised,zbontar2021barlow,grill2020bootstrap,chen2021exploring,chen2021empirical,caron2021emerging,zhou2022image,oquab2023dinov2}, and Masked Image Modeling (MIM), which learns to reconstruct missing pixels (or high-level representations) of images with occluded fragments~\cite{he2021masked,atito2021sit,xie2022simmim,bao2022beitbertpretrainingimage,peng2022beitv2maskedimage,assran2023ijepa}.
Although JEA representations often offer superior quality, they are highly dependent on the choice of data and pretraining augmentations, some of which may be detrimental to the performance of downstream tasks~\cite{tian2020makes,xiao2020whatshouldnotbecontrastive,przewiezlikowski2024augmentation,assran2023thehiddenuniform,oquab2023dinov2}.
In contrast, the advantage of MIM representations lies in a more generic pretext task that requires fewer assumptions about the pretraining data, thus increasing their applicability to non-standard data domains and downstream tasks~\cite{pardyl2023active,zhuang2024mimmaskmaskselfsupervised,chen2024probingmidlevelvisioncapabilities}. However, MIM representations often underperform in high-level perceptual tasks for reasons that are not fully understood~\cite{zhang2022howmask,park2023what,balestriero2024how}.

In this paper, we systematically analyze how masked models form their representations in order to understand the reasons for their poor quality. We find that MIM representations do not work well with the two standard ViT feature extraction methods -- the \cls{} tokens and average patch representations, which are commonly treated as global image descriptors~\cite{dosovitskiy2021image,caron2021emerging,he2021masked}. This is because, unlike JEAs, MIM representations are ineffective at aggregating the relevant semantic information (see left and center in \cref{fig:large_teaser}), which contributes to the performance gap between these two approaches.

These findings lead us to propose \textbf{Selective Aggregation} of MIM patch representations as a remedy. Using a lightweight technique inspired by Multiple-Instance Learning~\cite{ilse2018attention}, we consistently improve the quality of representation for a wide range of MIM models without fine-tuning their parameters (see~\cref{fig:teaser_barplot}). 
The improvements resulting from Selective Aggregation in the well-established~\cite{he2021masked,xie2022simmim} and recently published~\cite{assran2023ijepa,darcet2025clusterpredictlatentspatches} models support the key finding that the lack of proper aggregation is an inherent problem in MIMs. 
With the continued emergence of novel approaches~\cite{darcet2025clusterpredictlatentspatches}, we expect Selective Aggregation to remain a useful tool for their developers and users.

\textbf{Our contributions can be summarized as follows:}
\begin{itemize}
    \item We analyze the information flow within the widely used SSL models and show that MAE aggregates information from most image patches, while the competing approaches are more selective.
    \item We introduce Selective Aggregation of MIM patch tokens to properly extract their high-level information and thus consistently improve the performance of a wide variety of MIM models.
    \item We identify the lack of proper patch aggregation as an inherent problem in MIM, shedding new light on this SSL pre-training paradigm and providing important insights for its future development.
\end{itemize}

\section{Related works}

\paragraph{Self-supervised learning (SSL) of visual representations} 
has become a cornerstone of modern computer vision, enabling models to learn without labeled data~\citep{saleh2022selfsupervisedsurvey,balestriero2023cookbook}. 
Several powerful SSL paradigms have been developed, including Joint-Embedding Architectures (JEA)~\cite{he2020momentum,chen2020simple,caron2020unsupervised,caron2021emerging,oquab2023dinov2}, which learn representations by enforcing invariance across augmented image views, leading to strong out-of-the-box performance on high-level tasks.  
However, JEA approaches rely on carefully designed data augmentations~\cite{tian2020makes} and implicitly assume similar distributions between pretraining and downstream data~\cite{assran2023thehiddenuniform,oquab2023dinov2}, limiting their adaptability~\cite{xiao2020whatshouldnotbecontrastive,lee2021improving,chavhan2023amortised,przewiezlikowski2024augmentation,venkataramanan2024dora,garrido2024learningleveragingworldmodels}.  
As an alternative, Masked Image Modeling (MIM)~\cite{vincent2008extracting,vincent2010stacked,atito2021sit,he2021masked,xie2022simmim,darcet2025clusterpredictlatentspatches} reconstructs masked image regions or their representations, leveraging Transformers' ability to model long-range dependencies~\cite{he2021masked,peng2022beitv2maskedimage,pardyl2023active}. 
This paradigm has demonstrated strong fine-tuning performance and scalability~\cite{he2021masked,liu2024exploring,assran2023ijepa,ravi2024sam2segmentimages}, motivating further study into how MIM models structure information and how their representations can be effectively utilized~\cite{zhang2022howmask,park2023what,balestriero2024how}. 
Our work investigates this problem by analyzing how MIM models structure information and identifying a crucial shortcoming in their attention mechanisms.

\paragraph{Differences in representation structure between JEA and MIM} 
have been the subject of several studies analyzing their attention patterns and feature organization~\cite{zhang2022howmask,kong2023understanding,park2023what,balestriero2024how}. 
JEA models are known to produce compact, global representations, often relying on the \cls{} token to aggregate features~\cite{caron2021emerging,zbontar2021barlow}. 
In contrast, prior work has shown that MIM models tend to focus on local structure~\cite{park2023what,huang2024how,xie2023revealing}, leaving open the question of how their learned representations interact across tokens and how suitable they are for typical probing strategies in downstream tasks.  
Rather than directly addressing these differences, recent works propose to probe ViTs with additional attention layers~\cite{chen2023context,bardes2024vjepa} containing significantly more trainable parameters. However, the reason why such complex probing is needed remains unexplored.  
Our work fills this gap by systematically analyzing the information flow in ViTs pretrained with JEA and MIM, uncovering previously overlooked fundamental structural differences between both paradigms.  
Furthermore, we show that these differences contribute to inefficiencies when using MIMs for high-level perception tasks, highlighting the need for a lightweight probing approach that accounts for the lack of appropriate representation structure in MIMs.

\section{Preliminaries}

In this section, we recall the basic Vision Transformer (ViT) architecture~\citep{dosovitskiy2021image}, and the Masked Autoencoder (MAE)~\cite{he2021masked} -- the most popular Masked Image Modeling technique. 

\subsection{Vision transformers (ViT)}

\paragraph{Image processing by ViT} begins by dividing and flattening an image $\mathbf{x} \in \mathbb{R}^{H \times W \times C}$
into a sequence of $N$ non-overlapping \emph{patches} $\mathbf{x}_p \in \mathbb{R}^{N \times (P^2 \cdot C)}$, where $(P, P)$ is the resolution of a patch and $N = \frac{HW}{P^2}$. Next, a linear projection layer $e: \mathbb{R}^{(P^2 \cdot C)} \rightarrow  \mathbb{R}^D$ transforms each patch into a $D$-dimensional embedding to which appropriate positional encoding vectors $\mathbf{p}\in \mathbb{R}^{N \times D}$~\cite{dosovitskiy2021image} are added.
We refer to the result of these operations as \emph{patch tokens}:
\begin{align}
    \mathbf{z}_p = e(\mathbf{x}_p) + \mathbf{p} \in \mathbb{R}^{N \times D}. 
\end{align}
We also define a learnable \texttt{[cls]} token $\mathbf{x}_{cls} \in \mathbb{R}^D$, which is prepended to $\mathbf{z}_p$\footnote{For convenience of notation, the \texttt{[cls]} token will have the index of 0, and patch tokens will have the indices $\in 1...N$.}. 
The first ViT block input is defined as:
\begin{align}
        \mathbf{z}_0 = [\mathbf{x}_{cls}; \mathbf{z}_p] \in \mathbb{R}^{(N+1)\times D}
\end{align}
The $l$-th ViT block transforms tokens $\mathbf{z}_{l-1}$ into tokens $\mathbf{z}_l$. Each of the $L$ blocks is a sequence of Multihead Self-Attention (MSA)~\cite{vasvani2017attention} and MLP layers. For both MSA and MLP, the input is first normalized with LayerNorm~\cite{ba2016layernormalization}, and the output of the layer is summed with the unnormalized input, forming a residual connection~\cite{he2015deep}. 

\paragraph{Multihead Self-attention (MSA)}~\cite{vasvani2017attention} is a key component of ViT, which allows for exchanging image information between tokens. It consists of $h$ {self-attention heads}, each of which separately transforms the sequence of $(N+1)$ input tokens into a sequence of output tokens of the same length. 
A self-attention head creates three linear projections of the input, $\{\mathbf{q, k,v}\} \in \mathbb{R}^{(N+1)\times(D/h)}$ and computes the {self-attention map} $\mathbf{a} \in [0,1]^{(N+1) \times (N+1)}$:
\begin{align}
    \label{eq:selfattention}
    \mathbf{a} = softmax(\dfrac{\mathbf{q} \mathbf{k}^T}{\sqrt{D/h}}), 
\end{align}
Output tokens $\mathbf{o} \in \mathbb{R}^{(N+1)\times(D/h)}$ are calculated as $\mathbf{o} = \mathbf{a} \mathbf{v}$, i.e. the sums of $\mathbf{v}$ weighted by subsequent rows of $\mathbf{a}$.
Next, the output tokens of each self-attention head are concatenated along their token dimension and projected through a linear layer to form the final output of the MSA.

\paragraph{Final vision transformer representation} $\mathbf{z}_L$ consists of $(N+1)$ tokens of shape $D$. In high-level perception tasks such as image classification, the most common strategy is to use only the \cls{} token output of the final ViT block ($\mathbf{z}_{L, 0}$) as the representation of the entire image which serves as an input to the classifier~\citep{dosovitskiy2021image,caron2021emerging,zhang2022howmask}. The same approach is used in JEA pretraining, where the invariance objective is imposed on the \cls{} representations (typically followed by a projector network~\cite{chen2020simple,bordes2022guillotine}), while patch tokens are discarded~\cite{caron2021emerging,chen2021empirical}.
An alternative strategy is to summarize the image representation as the average value of patch tokens, i.e. $\sum_{i=1}^N \frac{\mathbf{z}_{L,i}}{N}$, sometimes even removing the \cls{} token from the model~\citep{he2021masked,assran2022masked}. However, this typically leads to representations of worse quality~\citep{dosovitskiy2021image}.

\subsection{Masked Image Modeling}

Masked Image Modeling (MIM)~\cite{vincent2008extracting,vincent2010stacked} is a paradigm of learning representations through the task of image inpainting (masking random contents of images and training a model to reconstruct them). This approach is straightforward to apply in vision transformers because masking can be implemented by randomly removing a subset of patch tokens. Among the various MIM implementations~\citep{xie2022simmim,atito2021sit}, the Masked Autoencoder (MAE)~\citep{he2021masked} has emerged as one of the most popular frameworks.

\paragraph{Masked Autoencoder (MAE)} consists of two ViTs -- an encoder $f$ and decoder $g$.
During MAE pretraining, we divide the image into patch tokens $\mathbf{z}_p$, remove a random subset of tokens, and then process the remaining ones through the encoder. The tokens to be removed are selected by a random binary mask $m \in \{0,1\}^N$, where $0$ is drawn with the probability of $\rho$ (mask ratio) and denotes the dropped tokens. In consequence, the input and output sequences of $f$ consist of $(1+ N\cdot (1-\rho))$ tokens (the \cls{} token and $N\cdot (1-\rho$) patch tokens).

Before processing the output of $f$ through the decoder\footnote{For simplicity of notation, we assume that the encoder and decoder have equal embedding sizes and numbers of layers, denoted by $D$ and $L$, respectively. In practice, if the embedding sizes are not equal, we prepend the decoder with an appropriate linear projection.} $g$, we complement it with $N \cdot \rho$ identical \emph{mask tokens} $\mathbf{z}_{msk} \in D$, such that the placement of mask tokens reflects the placement of tokens removed by mask $m$. The decoder adds an appropriate positional embedding to both, encoded and mask tokens.
After obtaining the output sequence of $g$, we discard the \cls{} token and project the $N$ patch tokens into the sequence $\hat{\mathbf{x}_p} \in \mathbb{R}^{N \times (P^2 \cdot C)}$, i.e. of the same size as the image patches $\mathbf{x}_p$.

The objective function of MAE is defined as the mean squared error between the image pixels and predicted pixels, calculated at the patches that were randomly dropped by mask $m$:
\begin{align}
     \mathcal{L}_{MAE} = \mathbb{E}_{\mathbf{x}}|| \mathbf{x}_p[1-m] - \hat{\mathbf{x}_p}[1-m] ||^2.
\end{align}

Numerous works propose to replace the MAE prediction target with higher-level representations of patches. Such targets can be formed from low-variance image components~\cite{wei2023maskedfeaturepredictionselfsupervised,balestriero2024how}, or latent representations of an image encoder~\cite{bao2022beitbertpretrainingimage,peng2022beitv2maskedimage,liu2024exploring,assran2023ijepa,darcet2025clusterpredictlatentspatches}. However, the reconstruction objective is typically applied to the mask tokens, whereas the \cls{} representation does not optimize any objective. 
This raises the question of what representation is formed by \cls{} token, and whether it is the optimal choice for a global descriptor in high-level perception tasks.

\section{Information flow in MIM and JEA}
\label{sec:attn_analysis}

The \cls{} token in Masked Image Models (MIMs) captures a representation that can, to some degree, serve as a global image descriptor~\cite{he2021masked,xie2022simmim}.  
However, its out-of-the-box quality is significantly lower than the \cls{} token obtained from Joint-Embedding Architectures (JEAs), limiting the effectiveness of standard probing techniques. 
This raises the question: \textit{What are the differences in how the \cls{} tokens gather information in these two approaches?}
Understanding these differences will allow us to build a deeper understanding of the MIM models and, in consequence, develop a principled approach to feature extraction.

In order to characterize the differences in the representational structure of vision transformers pretrained with MIM and JEA paradigms, we study their self-attention mechanism, as it is the only means by which the \cls{} token acquires information from the image patches.

\paragraph{Methodology.}

In self-attention, each token either recycles its representation by attending to itself or gathers the representations of other tokens by attending to them. We analyze these interactions to understand how information flows between \cls{} and patch tokens in publicly available ViTs pretrained with several popular SSL approaches~\cite{caron2021emerging,chen2021empirical,zhou2022image}, including the most popular MIM -- the Masked Autoencoder (MAE)~\cite{he2021masked}.\\ 

\vspace{2\baselineskip}
\noindent \textbf{Specifically, we measure:}  
\begin{itemize}
    \item \textbf{for the \cls{} token:} \begin{itemize}
        \item the proportion of attention the \cls{} token assigns to itself (\cref{fig:cls_cls_attention})
        \item the entropy of \cls{} attention to the patch tokens, quantifying the uniformity of attention distribution (\cref{fig:cls_pos_entropy})
    \end{itemize}
    \item \textbf{for each patch token:}
    \begin{itemize}
        \item the proportion of self-attention a token assigns to itself relative to its total attention to all patch tokens (\cref{fig:pos_self_attn_adj})
        \item the entropy of token attention to all patch tokens, measuring how selectively information is exchanged between patches (\cref{fig:pos_pos_entropy}).
    \end{itemize}
\end{itemize}
We provide the analysis for ViT-B models below and refer to \Cref{sec:app:info_flow} for a detailed methodology and the analysis conducted for ViT-S and ViT-L models.

\paragraph{Key findings.} 
Our analysis reveals significant differences in how information is exchanged between tokens of JEA- and MAE-trained ViTs.
The \cls{} token in JEA strongly attends to selective patch tokens, allowing it to integrate relevant information across ViT blocks. 
In contrast, the MAE \cls{} token heavily recycles its representation, limiting its ability to aggregate new information. Moreover, the remaining attention of the \cls{} token is almost uniformly distributed across all patch tokens, potentially absorbing redundant or irrelevant information. Crucially, fine-tuning MAE for classification shifts the attention of \cls{} and patches closer to that of JEA, highlighting the importance of selective attention in forming strong representations. In the following sections, we present our analysis in detail.

\subsection{Attention of the \cls{} token}

We observe significant differences in the behavior of \cls{} tokens between models pretrained with MAE and those pretrained with JEA methods, particularly in how they attend to themselves and to the patch tokens. We detail our study in the following paragraphs.

\begin{figure}[t]
    \centering
    \begin{tikzpicture}[
    node distance=1.25cm and 0.75cm,
]

    \node[draw] (zin0) {$\mathbf{v}_0$};
    \node[draw, right=1cm of zin0, gray] (zin1) {$\mathbf{v}_1$};
    \node[right=0.25cm of zin1, gray] (zindots) {$\dots$};
    \node[draw, right=0.25cm of zindots, gray] (zinN) {$\mathbf{v}_N$};

    \node[draw, below=1.7cm of zin0] (zout0) {$\mathbf{o}_0$};
    \node[right=of zout0] (zoutdots) {$\dots$};

    \draw[-stealth, thick] (zin0) -- (zout0) node[midway, fill=white, inner sep=3pt, draw=red, dashed, very thick, rounded corners] (a00) {$\mathbf{a}_{0,0}$};
    
    \draw[-stealth, thick, gray] (zin1.south) -- (zout0) node[midway, fill=white, inner sep=2pt] {$\mathbf{a}_{0,1}$};
    \draw[-stealth, thick, gray] (zinN.south) -- (zout0) node[midway, fill=white, inner sep=2pt] {$\mathbf{a}_{0,N}$};

    \draw[decorate,decoration={brace,amplitude=6pt}] 
        ([yshift=5pt]zin0.north west) -- ([yshift=5pt]zin0.north east) 
        node[midway, above=6pt] {\cls{} token};

    \draw[decorate,decoration={brace,amplitude=6pt},gray] 
        ([yshift=5pt]zin1.north west) -- ([yshift=5pt]zinN.north east) 
        node[midway, above=6pt] {Patch tokens};

    \node[left=of zin0, align=center] (int) {Value\\tokens};
    \node[below=of int, align=center] (outt) {Output\\tokens};
    \draw[-stealth, thick] (int) -- (outt) node[midway, fill=white, inner sep=1pt,align=center] {\scriptsize Attention};

\end{tikzpicture}

        \begin{tikzpicture}
        \begin{axis}[
            sec4style,
            ylabel={{\shortstack{Attention of the\\\cls{}token to itself}}},
	xlabel={ViT-B block}
	]
 \addplot[{DINOStyle}]
 table {%
1 0.1585644781589508
2 0.4371647536754608
3 0.4487747251987457
4 0.2436964064836502
5 0.08758749067783356
6 0.10344899445772171
7 0.1381075233221054
8 0.1890965849161148
9 0.18715742230415344
10 0.1650734394788742
11 0.1659216284751892
12 0.18133679032325745
};
\addlegendentry{DINO~\cite{caron2021emerging}}

 \addplot[{iBOTStyle}]
 table {%
1 0.17956870794296265
2 0.42291003465652466
3 0.6281848549842834
4 0.3125123083591461
5 0.14374788105487823
6 0.1439746916294098
7 0.10955562442541122
8 0.10184510052204132
9 0.1253819465637207
10 0.11884170770645142
11 0.11997197568416595
12 0.11452257633209229
};
\addlegendentry{iBOT~\cite{zhou2022image}}

 \addplot[{MoCoStyle}]
 table {%
1 0.3921937048435211
2 0.37001293897628784
3 0.5669093132019043
4 0.27546176314353943
5 0.19073139131069183
6 0.10868073999881744
7 0.11473498493432999
8 0.1351258009672165
9 0.09406556189060211
10 0.09086806327104568
11 0.05742160230875015
12 0.08490914106369019
};
\addlegendentry{MoCo-v3~\cite{chen2021empirical}}

\addplot[{MAEStyle}]
 table {%
1 0.4105907082557678
2 0.5290440917015076
3 0.40943580865859985
4 0.5888752341270447
5 0.540928304195404
6 0.7101895213127136
7 0.5992339849472046
8 0.6591408252716064
9 0.479407399892807
10 0.4309639036655426
11 0.5721790790557861
12 0.4142257571220398
};
\addlegendentry{MAE~\cite{he2021masked}}

 \addplot[{FinetunedMAEStyle}]
 table {%
1 0.6157293915748596
2 0.6317875981330872
3 0.6273791193962097
4 0.719139814376831
5 0.5765323042869568
6 0.6416050791740417
7 0.7379360795021057
8 0.7787177562713623
9 0.7339198589324951
10 0.710572361946106
11 0.6700039505958557
12 0.6385836005210876
};
\addlegendentry{MAE (FT)~\cite{he2021masked}}

\end{axis}
\end{tikzpicture}
    \caption{Attention of the \texttt{[cls]} token to itself is much higher in MAE, than in the JEA ViTs. As opposed to JEA, where the \cls{} tokens gather a large amount of information from the patch tokens, the MAE \cls{} tokens primarily recycles its own representation.
    }
    \label{fig:cls_cls_attention}
    \vspace{-1\baselineskip}
\end{figure}

\paragraph{The \cls{} token of MAE attends primarily to itself.}  
As shown in \cref{fig:cls_cls_attention}, the \cls{} token in MAE assigns a significantly higher proportion of attention to itself compared to JEA-trained ViTs. In contrast, JEA 
models gradually reduce self-attention in deeper blocks, allowing the \cls{} token to integrate more information from patch tokens. This suggests that MAE’s \cls{} token primarily recycles existing information rather than refining its representation through interaction with patches.
Surprisingly, fine-tuning MAE for classification increases its \cls{}-\cls{} self-attention even further. To gain further insight into this behavior and deepen our comparison between MIM and JEA, we next analyze how the \cls{} token distributes the remainder of its attention.

\paragraph{The \cls{} token of MAE attends to the patches too uniformly to select only the relevant ones.}
\begin{figure}[t]
    \centering
    \begin{tikzpicture}[
    node distance=1.25cm and 0.75cm,
]

    \node[draw, gray] (zin0) {$\mathbf{v}_0$};
    \node[draw, right=1 cm of zin0] (zin1) {$\mathbf{v}_1$};
    \node[right=of zin1] (zindots) {$\dots$};
    \node[draw, right=of zindots] (zinN) {$\mathbf{v}_N$};

    \node[draw, below=1.7 cm of zin0] (zout0) {$\mathbf{o}_0$};
    \node[right=of zout0] (zoutdots) {$\dots$};

    \draw[-stealth, thick, gray] (zin0) -- (zout0) node[midway, fill=white, inner sep=0pt] {$\mathbf{a}_{0,0}$};
    \draw[-stealth, thick] (zin1) -- (zout0) node[midway, fill=white, inner sep=2pt] (a01){$\mathbf{a}_{0,1}$};
    \draw[-stealth, thick] (zinN) -- (zout0) node[midway, fill=white, inner sep=2pt] (a0N) {$\mathbf{a}_{0,N}$};
    \node[left=0.01cm of a0N] {$\dots$};

    \draw[decorate,decoration={brace,amplitude=6pt}, gray] 
        ([yshift=5pt]zin0.north west) -- ([yshift=5pt]zin0.north east) 
        node[midway, above=6pt] {\cls{} token};

    \draw[decorate,decoration={brace,amplitude=6pt}] 
        ([yshift=5pt]zin1.north west) -- ([yshift=5pt]zinN.north east) 
        node[midway, above=6pt] {Patch tokens};

    \node[left=of zin0, align=center] (int) {Value\\tokens};
    \node[below=of int, align=center] (outt) {Output\\tokens};
    \draw[-stealth, thick] (int) -- (outt) node[midway, fill=white, inner sep=1pt,align=center] {\scriptsize Attention};

    \node[draw=red, dashed,fit=(a01) (a0N), very thick, inner sep=1pt, rounded corners] (a_container) {};

    \node[below right=0.1cm and 0.5cm of a_container, red] (ent_label) {Entropy}; %
    
    \draw[-stealth, very thick, dashed, red] (a_container) -- (ent_label); %
    
\end{tikzpicture}
        \begin{tikzpicture}
        \begin{axis}[
            sec4style,
            ylabel={{\shortstack{Entropy of attention between\\the \cls{}and patch tokens}}},
	xlabel={ViT-B block}
	]
 \addplot[{DINOStyle}]
 table {%
1 5.142787933349609
2 4.672569274902344
3 4.999399185180664
4 5.003344535827637
5 4.872176170349121
6 4.846645832061768
7 4.68649959564209
8 4.586240768432617
9 4.39156436920166
10 4.322977542877197
11 4.356078147888184
12 4.561920642852783
};
\addlegendentry{DINO~\cite{caron2021emerging}}

 \addplot[{iBOTStyle}]
 table {%
1 5.136229515075684
2 4.9121623039245605
3 4.688700199127197
4 5.02535343170166
5 4.955214500427246
6 4.915258884429932
7 4.813562393188477
8 4.644126892089844
9 4.465449333190918
10 4.3610382080078125
11 4.5107502937316895
12 4.679174423217773
};
\addlegendentry{iBOT~\cite{zhou2022image}}

 \addplot[{MoCoStyle}]
 table {%
1 4.251352787017822
2 4.919338703155518
3 4.876862525939941
4 5.0949177742004395
5 5.020232200622559
6 4.991518974304199
7 4.926223278045654
8 4.838733673095703
9 4.690496444702148
10 4.553460597991943
11 4.596993446350098
12 4.623865127563477
};
\addlegendentry{MoCo-v3~\cite{chen2021empirical}}

 \addplot[{MAEStyle}]
 table {%
1 5.036749362945557
2 4.80377197265625
3 4.993056774139404
4 5.023566722869873
5 5.124216556549072
6 5.077910423278809
7 5.099809646606445
8 5.040615081787109
9 5.039320468902588
10 5.027137279510498
11 5.051735877990723
12 5.025387287139893
};
\addlegendentry{MAE~\cite{he2021masked}}

 \addplot[{FinetunedMAEStyle}]
 table {%
1 5.05247163772583
2 4.924233436584473
3 4.977036476135254
4 4.975040435791016
5 5.07895565032959
6 5.055420875549316
7 4.958649635314941
8 4.9717698097229
9 4.844841003417969
10 4.596076488494873
11 4.419389247894287
12 4.2843475341796875
};
\addlegendentry{MAE (FT)~\cite{he2021masked}}

\end{axis}
\end{tikzpicture}
    \caption{Entropy of \cls{} token attention to patch tokens reaches almost the maximal possible level in MAE. In other models, it decreases in the deeper model blocks, indicating that the \cls{} token attends to different patches in a more selective manner. Fine-tuning of MAE decreases this entropy, indicating that selective attention to patch tokens is crucial for good perception.
    }
    \label{fig:cls_pos_entropy}
\end{figure}

\cref{fig:cls_pos_entropy} shows the entropy of attention between the \cls{} and patch tokens.  
In MAE, this entropy remains high throughout the ViT blocks, approaching its theoretical upper bound (5.27 for a discrete
distribution over 196 patches), indicating that \cls{} spreads its attention broadly rather than selectively attending to relevant patches.  
In contrast, JEA 
models exhibit lower entropy, meaning their \cls{} tokens focus on fewer, more important patches.  
Fine-tuning the MAE significantly reduces entropy, making its attention patterns more similar to JEA models.  
Furthermore, we hypothesize that as fine-tuning reduces attention to less relevant patches, the \cls{} token redistributes this attention toward itself, accounting for the increase in \cls{}-\cls{} attention observed in \cref{fig:cls_cls_attention}.  

Given that the \cls{} representations in joint-embedding ViTs and fine-tuned MAEs are much better suited for perception compared to their MAE counterparts, we hypothesize that their ability to selectively attend to relevant patch tokens is essential for forming high-quality global representations in ViTs -- yet this property does not naturally emerge in the MAE framework.

\begin{figure}[t]
    \centering
    \begin{tikzpicture}[
    node distance=1.25cm and 0.75cm,
]

    \node[draw, gray] (zin0) {$\mathbf{v}_0$};
    \node[draw, right=1 cm of zin0] (zin1) {$\mathbf{v}_1$};
    \node[right=0.1cm of zin1] (zindots) {$\dots$};
    \node[draw, right=0.1 cm of zindots] (zini) {$\mathbf{v}_i$};
    \node[right=0.1 cm of zini] (zindots2) {$\dots$};

    \node[draw, right=0.1 cm of zindots2] (zinN) {$\mathbf{v}_N$};

    \draw[decorate,decoration={brace,amplitude=6pt}, gray] 
        ([yshift=5pt]zin0.north west) -- ([yshift=5pt]zin0.north east) 
        node[midway, above=6pt] {\cls{} token};

    \draw[decorate,decoration={brace,amplitude=6pt}] 
        ([yshift=5pt]zin1.north west) -- ([yshift=5pt]zinN.north east) 
        node[midway, above=6pt] {Patch tokens};

    \node[left=of zin0, align=center] (int) {Value\\tokens};
    \node[below=of int, align=center] (outt) {Output\\tokens};
    \draw[-stealth, thick] (int) -- (outt) node[midway, fill=white, inner sep=1pt,align=center] {\scriptsize Attention};

    \node[draw, below=1.7 cm of zini] (zouti) {$\mathbf{o}_{l}^i$};

    \draw[-stealth, thick, gray] (zin0.south) -- (zouti) node[midway, fill=white, inner sep=0pt] {$\mathbf{a}_{i,0}$};
    \draw[-stealth, thick] (zin1) -- (zouti) node[midway, fill=white, inner sep=2pt] (ai1) {$\mathbf{a}_{i,1}$};
    \draw[-stealth, thick] (zini) -- (zouti) node[midway, fill=white, inner sep=3pt, draw=red, very thick, dashed, rounded corners] (aii) {$\mathbf{a}_{i,i}$};

    \draw[-stealth, thick] (zinN) -- (zouti) node[midway, fill=white, inner sep=2pt] (aiN) {$\mathbf{a}_{i,N}$};

    \node[draw=blue, dashed,fit=(ai1) (aiN), very thick, inner sep=3pt, rounded corners] (a_container) {};

    \node[below left= 0.4 cm and 0 cm of a_container, inner sep=-1.5pt] (ratio) {$\dfrac{\textcolor{red}{\mathlarger{\mathlarger{\bullet}}}}{\Sigma \textcolor{blue}{\mathlarger{\mathlarger{\bullet}}}}$};
    \draw[-stealth, very thick, dashed, red] (aii) -- (ratio.north east);
    \draw[-stealth, very thick, dashed, blue] (a_container) -- (ratio.south east);

    \node[below =1.9 cm of zindots2] {$\dots$};

\end{tikzpicture}     \begin{tikzpicture}
        \begin{axis}[
            sec4style,
            ylabel={{\shortstack{Relative attention between\\the same patch tokens}}},
	xlabel={ViT-B block}
	]
 \addplot[{DINOStyle}]
 table {%
1 0.10427000373601913
2 0.058244701474905014
3 0.04074423387646675
4 0.03401390463113785
5 0.031960275024175644
6 0.021716957911849022
7 0.019277183338999748
8 0.018366316333413124
9 0.011647745966911316
10 0.011410688050091267
11 0.009314093738794327
12 0.011983493342995644
};
\addlegendentry{DINO~\cite{caron2021emerging}}

 \addplot[{iBOTStyle}]
 table {%
1 0.08299794793128967
2 0.08324775099754333
3 0.06118907406926155
4 0.03964445739984512
5 0.04465824365615845
6 0.029700549319386482
7 0.0255900751799345
8 0.023498745635151863
9 0.017476389184594154
10 0.012934492900967598
11 0.02271910198032856
12 0.026191875338554382
};
\addlegendentry{iBOT~\cite{zhou2022image}}

 \addplot[{MoCoStyle}]
 table {%
1 0.012961355037987232
2 0.07047972828149796
3 0.04581468924880028
4 0.05335165560245514
5 0.053382404148578644
6 0.03896085172891617
7 0.03106638975441456
8 0.03525640442967415
9 0.01549577433615923
10 0.014685172587633133
11 0.008616182953119278
12 0.007575439289212227
};
\addlegendentry{MoCo-v3~\cite{chen2021empirical}}

\addplot[{MAEStyle}]
 table {%
1 0.030649458989501
2 0.07072880864143372
3 0.04010970890522003
4 0.033401232212781906
5 0.09110450744628906
6 0.049564212560653687
7 0.10953568667173386
8 0.07168903201818466
9 0.030591905117034912
10 0.03453614562749863
11 0.025126393884420395
12 0.025260325521230698
};
\addlegendentry{MAE~\cite{he2021masked}}

 \addplot[{FinetunedMAEStyle}]
 table {%
1 0.038292743265628815
2 0.08017653971910477
3 0.04946440830826759
4 0.045961491763591766
5 0.08078787475824356
6 0.03815276175737381
7 0.036439187824726105
8 0.02027437463402748
9 0.01154826395213604
10 0.0073655713349580765
11 0.0151986638084054
12 0.016380617395043373
};
\addlegendentry{MAE (FT)~\cite{he2021masked}}

\end{axis}
\end{tikzpicture}
    \caption{
    Attention of the patch tokens to themselves, relative to the total attention given to all patch tokens. In the later MAE blocks, patch tokens seem to allocate more relative attention to themselves, compared to JEA.}
    \label{fig:pos_self_attn_adj}
\end{figure}

\begin{figure}[t]
    \begin{tikzpicture}[
    node distance=1.25cm and 0.75cm,
]

    \node[draw, gray] (zin0) {$\mathbf{v}_0$};
    \node[draw, right=1 cm of zin0] (zin1) {$\mathbf{v}_1$};
    \node[right=0.1cm of zin1] (zindots) {$\dots$};
    \node[draw, right=0.1 cm of zindots] (zini) {$\mathbf{v}_i$};
    \node[right=0.1 cm of zini] (zindots2) {$\dots$};

    \node[draw, right=0.1 cm of zindots2] (zinN) {$\mathbf{v}_N$};

    \draw[decorate,decoration={brace,amplitude=6pt}, gray] 
        ([yshift=5pt]zin0.north west) -- ([yshift=5pt]zin0.north east) 
        node[midway, above=6pt] {\cls{} token};

    \draw[decorate,decoration={brace,amplitude=6pt}] 
        ([yshift=5pt]zin1.north west) -- ([yshift=5pt]zinN.north east) 
        node[midway, above=6pt] {Patch tokens};

    \node[left=of zin0, align=center] (int) {Value\\tokens};
    \node[below=of int, align=center] (outt) {Output\\tokens};
    \draw[-stealth, thick] (int) -- (outt) node[midway, fill=white, inner sep=1pt,align=center] {\scriptsize Attention};

    \node[draw, below=1.7 cm of zini] (zouti) {$\mathbf{o}_i$};

    \draw[-stealth, thick, gray] (zin0.south) -- (zouti) node[midway, fill=white, inner sep=0pt] {$\mathbf{a}_{i,0}$};
    \draw[-stealth, thick] (zin1) -- (zouti) node[midway, fill=white, inner sep=2pt] (ai1) {$\mathbf{a}_{i,1}$};
    \draw[-stealth, thick] (zini) -- (zouti) node[midway, fill=white, inner sep=2pt] (aii) {$\mathbf{a}_{i,i}$};

    \draw[-stealth, thick] (zinN) -- (zouti) node[midway, fill=white, inner sep=2pt] (aiN) {$\mathbf{a}_{i,N}$};

    \node[draw=red, dashed,fit=(ai1) (aiN), very thick, inner sep=2pt, rounded corners] (a_container) {};

    \node[below right= 0.25 cm and -0.4 cm of a_container, red] (label) {Entropy};
    \draw[-stealth, very thick, dashed, red] (a_container) -- (label);

    \node[below=1.85cm of zindots] {$\dots$};

\end{tikzpicture}
        \begin{tikzpicture}
        \begin{axis}[
            sec4style,
            ylabel={{\shortstack{Entropy of attention\\between patch tokens}}},
	xlabel={ViT-B block}
	]
 \addplot[{DINOStyle}]
 table {%
1 4.3837690353393555
2 3.6690053939819336
3 3.6203320026397705
4 3.6401712894439697
5 3.7775065898895264
6 4.022028923034668
7 4.013164043426514
8 4.112133979797363
9 4.503214359283447
10 4.712254524230957
11 4.827049732208252
12 4.955743312835693
};
\addlegendentry{DINO~\cite{caron2021emerging}}

 \addplot[{iBOTStyle}]
 table {%
1 4.466940879821777
2 3.658950090408325
3 3.2001729011535645
4 3.753647565841675
5 3.8279638290405273
6 4.036083698272705
7 4.13369083404541
8 4.075435638427734
9 4.273280143737793
10 4.579384803771973
11 4.731393814086914
12 4.682201862335205
};
\addlegendentry{iBOT~\cite{zhou2022image}}

 \addplot[{MoCoStyle}]
 table {%
1 4.0157294273376465
2 4.436943054199219
3 3.4602842330932617
4 3.637147903442383
5 3.8815367221832275
6 4.102165222167969
7 4.23651123046875
8 4.000644207000732
9 4.430052757263184
10 4.479116916656494
11 4.757903575897217
12 5.199434757232666
};
\addlegendentry{MoCo-v3~\cite{chen2021empirical}}

 \addplot[{MAEStyle}]
 table {%
1 4.718160629272461
2 4.302433490753174
3 4.403627395629883
4 4.527017593383789
5 2.737722873687744
6 3.0187857151031494
7 2.9202182292938232
8 3.9761245250701904
9 4.13281774520874
10 4.054203510284424
11 4.30564022064209
12 4.246712684631348
};
\addlegendentry{MAE~\cite{he2021masked}}

 \addplot[{FinetunedMAEStyle}]
 table {%
1 4.532347679138184
2 4.233613967895508
3 4.2133469581604
4 4.285141468048096
5 2.9295084476470947
6 3.644650459289551
7 3.7847113609313965
8 4.129693031311035
9 4.158210754394531
10 4.082627773284912
11 3.9730658531188965
12 3.981553316116333
};
\addlegendentry{MAE (FT)~\cite{he2021masked}}

\end{axis}
\end{tikzpicture}
    \caption{Entropy of patch tokens attention to themselves. In MAE, the patch tokens attend to other patches with lower entropy than in JEA,  indicating that they form a representation of local image fragments.
    }
    \label{fig:pos_pos_entropy}
\end{figure}

\subsection{Attention of the patch tokens}
We next analyze how patch tokens exchange information by measuring their self-attention (relative to total patch attention) and the entropy of their attention distribution across patches.

\paragraph{The patch tokens of MAE assign more attention to themselves.}

\cref{fig:pos_self_attn_adj} shows that patch tokens in MAE self-attend more than those in JEA 
models. This suggests that MAE patches prioritize local information over exchanging content with other patches, reinforcing their role in capturing fine-grained localized image details~\cite{park2023what}.

\paragraph{Patch tokens of MAE attend to patches more selectively than those of JEA.}  
The above findings are further reinforced by \cref{fig:pos_pos_entropy}, which shows that MAE patch tokens attend to other patches with lower entropy than those in JEA models, suggesting more localized and selective information exchange. This aligns with prior findings that MAE patches form semantically meaningful clusters~\cite{shin2024selfguided} and exhibit sparse, localized attention compared to JEA, where patch attention is more homogeneous~\cite{park2023what}. These results indicate that MAE patch tokens capture diverse, detailed local representations, but exchange less information across the image.

\section{Selective Aggregation of Masked Image Modeling representations}

Our analysis showed that masked models do not form structured global representations as effectively as JEA models because their \cls{} tokens do not properly aggregate high-level information from the relevant patches. Instead, they spread attention broadly, absorbing both relevant and redundant content. Patch averaging is an alternative, but it treats all patches equally and fails to prioritize the most informative ones. This leads us to ask: 
\textit{Can we improve the quality of the MIM representation simply by modifying its aggregation scheme?}

To address this, we propose \textbf{Selective Aggregation}, a mechanism that dynamically assigns importance to tokens when forming the final representation. 
Specifically, we define an aggregation function ${s: \mathbb{R}^{N \times D} \rightarrow [0, 1]^{N}}$ that predicts a score vector $\mathbf{s} \in [0,1]^{N+1}$ weighting patch tokens from the $L$-th ViT encoder block $\mathbf{z}_{L,1:N} \in \mathbb{R}^{N \times D}$ in a summation-based aggregation mechanism~\cite{bahdanau2015neural}.
The weights of $\mathbf{s}$ identify the key tokens and aggregate them into the representation $\mathbf{z}_{\text{select}} = \sum\limits_{i=0}^{N} \mathbf{s}_i \mathbf{z}_{L,i} \in \mathbb{R}^D$, which can then be used as a drop-in replacement for the \cls{} token or the naively averaged representation.
The existence of a function $s$ that aggregates tokens into a representation better than the \cls{} token would indicate that the MIM patch tokens actually contain high-level information that has not been captured by \cls{}, supporting our hypothesis that MIM models do not naturally form structured global representations.

We implement Selective Aggregation with Attention-based Multiple Instance Learning Pooling (AbMILP)~\cite{ilse2018attention} -- an approach that dynamically assigns importance weights to tokens, enabling structured aggregation while maintaining minimal complexity.  
Given a set of vectors (in our case, tokens $\mathbf{z}_{L}$), AbMILP predicts aggregation weights by applying a linear model $t: \mathbb{R}^{D} \rightarrow \mathbb{R}$ to each vector, followed by softmax:  
\begin{align}
    \mathbf{s}^{\text{AbMILP}}_i = \frac{ \exp(t(\mathbf{z}_{L,i}))} {\sum\limits_{j=0}^N \exp (t(\mathbf{z}_{L,j}))}.
    \label{eq:abmilp}
\end{align}

Crucially, Selective Aggregation only restructures the existing out-of-the-box ViT representations without transforming them into a different representation space. 
This ensures that our evaluation isolates the impact of aggregation itself, without modifying confounding factors such as the inherent quality of MIM token representations~\cite{balestriero2024how,alkin2025mimrefiner}.  
From a practical standpoint, this allows for a lightweight implementation of the aggregation function.

In the following sections, we evaluate the high-level representations of MIMs equipped with Selective Aggregation, and discuss the practical aspects of the aggregation mechanism. We ablate the design of the $t$ function in AbMILP and explore alternative aggregation functions in~\cref{sec:ablation}. In~\cref{app:sec:localization}, we discuss the use of Selective Aggregation scores for object localization.

\subsection{Evaluation of Selective Aggregation in high-level perception tasks}
\label{sec:main_experiments}

We evaluate how Selective Aggregation affects the global representations of vision transformers in several downstream tasks, including ImageNet-1k classification~\cite{russakovsky2015imagenet}, few-shot classification (ImageNet-1\%~\cite{assran2022masked,assran2023ijepa}) and fine-grained recognition (CUB-200~\cite{wah2011cub}).
\paragraph{Our evaluation follows several principles:}
\begin{itemize}
    \item We evaluate a wide range of prominent SSL ViTs using parameters made publicly available by their authors~\cite{he2021masked,zhuang2024mimmaskmaskselfsupervised,xie2022simmim,peng2022beitv2maskedimage,assran2023ijepa,oquab2023dinov2,darcet2025clusterpredictlatentspatches,chen2021empirical,zhou2022image,dosovitskiy2021image}\footnote{Due to the lack of publicly available parameters of ViT-S trained with MAE, we train this model with the same procedure as ViT-B~\cite{he2021masked}.}. Except for DINO-v2~\cite{oquab2023dinov2}, all models are pretrained on ImageNet-1k\footnote{For BEIT-v2~\cite{peng2022beitv2maskedimage}, we use the variant of the encoder without the intermediate ImageNet-21k finetuning.}.
    \item We do not fine-tune the parameters of evaluated models, but only train the classification heads that use their out-of-the-box representations. The AbMILP module is trained jointly with the classification head.
    \item We do not use techniques improving the linear probing performance, such as combining representations from ViT blocks other than the last one~\cite{caron2021emerging,peng2022beitv2maskedimage}\footnote{When using the SimMIM parameters, we use the representations from the 8-th ViT block, as recommended by the authors~\cite{xie2022simmim}.}.
    \item The hyperparameters of our evaluation follow the MAE linear probing protocol~\cite{he2021masked} and are described in detail in~\cref{app:sec:hyperparams}.
\end{itemize}

\begin{table}[t]
    \centering
      \resizebox{\linewidth}{!}{
    \begin{tabular}{ccccccc}
        \toprule
        \multicolumn{3}{c}{\textbf{Encoder}} & \multicolumn{4}{c}{\textbf{Representation aggregation method}} \\
        \multicolumn{2}{c}{\multirow{2}{*}{Source}} & \multirow{2}{*}{ViT}&  Avg. pooling & \cls  & \multicolumn{2}{c}{\emph{Selective (ours)}} \\ 
        & & & of patches & token & \emph{patches} & \emph{+ \cls{}} \\
         \midrule \midrule
        \multirow{9}{*}{\rotatebox[origin=c]{90}{\textbf{Masked Image Modeling}}} & 
        MAE~\cite{he2021masked} & ViT-S &  47.1 & 47.4 & \textbf{54.4} & \textbf{54.6} \\
        & MAE~\cite{he2021masked} & ViT-B &  65.8 & 67.8 & \textbf{71.6} &  \textbf{71.5} \\
        & MAE~\cite{he2021masked} & ViT-L & 73.0 & 75.8 &  \textbf{77.4} &  \textbf{77.4} \\
        & MAE~\cite{he2021masked} & ViT-H &  73.8 & 77.0 & \textbf{78.1} & \textbf{78.0} \\
        \cmidrule{2-7}
        & SimMIM~\cite{xie2022simmim} & ViT-B &  54.3 & 51.5 & \textbf{62.8} &  {62.0} \\
        & MaskFeat~\cite{wei2023maskedfeaturepredictionselfsupervised} & ViT-B &  56.9 & 62.9 & \textbf{66.6} & 65.8 \\
        & BEIT-v2~\cite{peng2022beitv2maskedimage} & ViT-B & 78.5 & 78.9 & \textbf{80.9} & \textbf{81.0} \\
        & I-JEPA~\cite{assran2023ijepa} & ViT-H &  77.7 & -- &\textbf{79.2} & - \\
        & CAPI~\cite{darcet2025clusterpredictlatentspatches} & ViT-L & {76.2} & -- & \textbf{82.4} & - \\
        \midrule
        \multirow{4}{*}{\rotatebox[origin=c]{90}{\textbf{JEA}}}& iBOT~\cite{zhou2022image} & ViT-B  & 75.0 & {77.8}& {77.9} & \textbf{78.2} \\
        & {DINO-v2}~\cite{oquab2023dinov2} & ViT-B & 81.9 & 83.2  & \textbf{83.5} & \textbf{83.5} \\
        & DINO~\cite{caron2021emerging} &  ViT-B &  71.1 & \textbf{76.6} &  75.2 & 76.2 \\
        & MoCo-v3~\cite{chen2021empirical} & ViT-B &  71.1 & \textbf{75.1} & \textbf{75.1} & \textbf{75.2} \\
        \midrule
        & MAE (+ FT)~\cite{he2021masked} & ViT-B &  76.6 & \textbf{80.0} & {79.1} & \textbf{79.8} \\
        \bottomrule
    \end{tabular}
    }
    \caption{
    Linear probing accuracy on ImageNet-1k~\cite{russakovsky2015imagenet} for different global image representations. In Masked Image Models, patch tokens aggregated via Selective Aggregation consistently produce global representations of higher quality than those obtained from the \cls{} and naively averaged patch tokens. 
    }
    \label{tab:main_results}
\end{table}

\begin{table*}[t]
    \centering
      \resizebox{\linewidth}{!}{
    \begin{tabular}{cccc|cc|cc|cc|cc|cc|cc}
        \toprule
       \multicolumn{2}{c}{\textbf{Encoder}} & \multicolumn{2}{c|}{\textbf{ImageNet-1\%}~\cite{russakovsky2015imagenet}} & \multicolumn{2}{c|}{\textbf{CUB}~\cite{wah2011cub}} & \multicolumn{2}{c|}{\textbf{Stanford Cars}~\cite{krause2013cars}} & \multicolumn{2}{c|}{\textbf{OxfordIIIPets}~\cite{parkhi2012pets}}  & \multicolumn{2}{c|}{\textbf{Food-101}~\cite{bossard2014food101}}  & \multicolumn{2}{c|}{\textbf{NUS-WIDE}~\cite{chua2009nuswide}}  & \multicolumn{2}{c}{\textbf{COCO}~\cite{tsung2014cocodataset}} \\
       {Source} & {ViT} &  Standard  & \emph{Selective} & Standard &\emph{Selective} & Standard &\emph{Selective} & Standard &\emph{Selective} & Standard &\emph{Selective} & Standard &\emph{Selective} & Standard &\emph{Selective} \\
         \midrule \midrule
        MAE~\cite{he2021masked} & ViT-B & 39.1 & \textbf{48.3} & 45.8 & \textbf{65.9} & 31.8 & \textbf{58.8} & 82.4 & \textbf{90.8} & 68.1 & \textbf{78.0} & 67.2 & \textbf{67.9} & 61.2 & \textbf{65.3}  \\
        SimMIM~\cite{xie2022simmim} & ViT-B & 17.3 & \textbf{34.5} & 17.9 & \textbf{61.8} & 11.3 & \textbf{23.5} & 37.9 & \textbf{47.7} & 54.2 & \textbf{61.9} & 58.8 & \textbf{60.0} & 41.5 & \textbf{44.9} \\
        BEIT-v2~\cite{peng2022beitv2maskedimage} & ViT-B & 66.8 & \textbf{69.0} & 79.2 & \textbf{80.4} & 72.2 & \textbf{74.9} & \textbf{93.7} & 93.4 & 88.2 & \textbf{90.2} & 69.5 & \textbf{71.9} & 71.4 & \textbf{76.9} \\
        I-JEPA~\cite{assran2023ijepa} & ViT-H & 66.4 & \textbf{70.9} & 51.7 & \textbf{59.9}  & 40.9 & \textbf{42.7} & 89.9 & \textbf{92.4} & 81.1 & \textbf{85.1} & 71.7 & \textbf{72.2} & 70.7 & \textbf{73.6} \\
        CAPI~\cite{darcet2025clusterpredictlatentspatches} & ViT-L & 52.7 & \textbf{74.2} & 25.9 & \textbf{79.7}  & 45.6 & \textbf{76.8} & 83.8 & \textbf{94.5} & 85.9 & \textbf{90.5} & 71.8 & \textbf{73.3} & 72.3 & \textbf{80.1}  \\
        \bottomrule
    \end{tabular}
    }
    \caption{Evaluation of standard (\cls{} for all models, except for I-JEPA~\cite{assran2023ijepa} and CAPI~\cite{darcet2025clusterpredictlatentspatches}), and selectively aggregated MIM representations on low-shot (ImageNet-1\%~\cite{russakovsky2015imagenet}), fine-grained (CUB~\cite{wah2011cub}, Stanford Cars~\cite{krause2013cars}, OxfordIIIPets~\cite{parkhi2012pets}, Food-101~\cite{bossard2014food101}), and multilabel (NUS-WIDE~\cite{chua2009nuswide}, COCO~\cite{tsung2014cocodataset}) classification tasks. Selective Aggregation consistently improves or matches MIM performance on these tasks.
    }  
    \label{tab:lowshot}
    \vspace{-0.5cm}
\end{table*}

\paragraph{ImageNet-1k classification (\cref{tab:main_results}).}
We evaluate the quality of representations formed by the \cls{} token, average patch representation, and Selective Aggregation. To understand the effect of Selective Aggregation, we apply it to a wide selection of prominent MIM and JEA models in two variants: \textbf{(i)} aggregating only the patch tokens, and \textbf{(ii)} aggregating the patch and the \cls{} tokens\footnote{I-JEPA~\cite{assran2023ijepa} and CAPI~\cite{darcet2025clusterpredictlatentspatches} do not include the \cls{} tokens in their architecture.}. The key takeaways are summarized below:
\begin{itemize}
    \item \textbf{Selective Aggregation consistently benefits Masked Image Models.} We observe consistent improvements in a wide variety of MIMs which were pretrained with both low-level~\cite{he2021masked, xie2022simmim, wei2023maskedfeaturepredictionselfsupervised}), and high-level~\cite{peng2022beitv2maskedimage,assran2023ijepa,darcet2025clusterpredictlatentspatches} prediction targets. This supports our hypothesis that the lack of such aggregation is an inherent problem in MIMs, regardless of how they are trained.
    \item \textbf{JEAs do not require Selective Aggregation.} In JEAs, Selective Aggregation and the~\cls{} token representations have similar quality, confirming that these models can be used out-of-the-box to select relevant patches.
    A slight improvement can be observed in iBOT~\cite{zhou2022image} and DINO-v2~\cite{oquab2023dinov2}, which use mask modeling of their own patch representations as a secondary training objective to JEA.
    \item \textbf{Aggregating the \cls{} is insignificant.} Aggregating the \cls{} token with patches is insignificant in MIMs, further confirming its low representation quality. In JEAs, it tends to improve the results because their \cls{} tokens already contain rich representations.
\end{itemize}

\paragraph{Low-shot and fine-grained classification (\cref{tab:lowshot}).} Having established that Selective Aggregation improves MIM performance, we further evaluate it with several MIM models on the more challenging low-shot, fine-grained, and multilabel classification tasks. The favorable performance of Selective Aggregation further reinforces its usefulness.

\subsection{Overhead of Selective Aggregation} 
The AbMILP-based token aggregator consists of a lightweight linear regressor that maps the representation vectors of dimension $D$ to scalars (i.e. the model $t$ in \cref{eq:abmilp}). 
At the same time, the classifier is a single linear layer that maps the representation vectors of dimension $D$ to logits of dimension $K$, equal to the number of classes (in our case, $K=1,000$).
As a result, the number of trainable parameters increases slightly, from $(D + 1) \cdot K$ to $(D +1) \cdot (K+1)$, with negligible computational and parameter overhead.

\section{Conclusion}

Masked Image Models (MIMs) are increasingly popular, yet their out-of-the-box usefulness in high-level perception tasks is suboptimal. This paper presents an in-depth analysis of why that is the case. We analyze the attention of \cls{} token for various SSL approaches and conclude that MIMs attend more uniformly to all patches when producing global representation. In contrast, better-performing Joint-Embedding Architectures (JEAs) are more selective and, as a result, accumulate only relevant information.

As a remedy, we propose Selective Aggregation of the patch representations returned by MIM. We demonstrate that this approach consistently improves the perception performance of multiple MIM models, regardless of whether their original prediction target was low-level pixels or high-level latent representations. 

These results support the hypothesis that a proper aggregation of the information stored in the patch tokens is crucial for high-quality representations in vision transformers.
We hope that this new perspective on MIM representations will inspire future work on improving these models, and pave the way for their broader practical applications.

\paragraph{Limitations.} Our analysis is based on models pretrained by the original authors, which limits our ability to explore model variations or hyperparameter choices, as only a single configuration was provided. Additionally, we have not tested all possible variants of JEA and MIM models, so our findings may not generalize to all configurations.

\paragraph{Impact statement.} This work advances our understanding of self-supervised vision transformers and opens new avenues for improving MIM models. By highlighting the importance of Selective Aggregation, it paves the way for future research focused on developing more efficient and effective self-supervised learning techniques, with the potential to significantly advance high-level perception tasks.

\section*{Acknowledgements}

This research has been supported by the flagship project entitled “Artificial Intelligence Computing Center Core Facility” from the Priority Research Area DigiWorld under the Strategic Programme Excellence Initiative at the Jagiellonian University.
The research of Marcin Przewięźlikowski was supported by the National Science Centre (Poland), grant no. 2023/49/N/ST6/03268.  
The research of Wojciech Jasiński was supported by the National Science Centre (Poland), grant no. 2022/47/B/ST6/03397. 
The research of Marek Śmieja was supported by the National Science Centre (Poland), grant no. 2022/45/B/ST6/01117.
The research of Bartosz Zieliński was supported by the National Science Centre (Poland), grant no. 2023/50/E/ST6/00469.
We gratefully acknowledge Polish high-performance computing infrastructure PLGrid (HPC Center: ACK Cyfronet AGH) for providing computer facilities and support within computational grant no. PLG/2024/017148.
We thank Marcin Sendera, Adam Pardyl, Turhan Can Kargin, Bill Psomas, Michał Pietruszka, and Klaudia Bałazy for fruitful discussions and feedback over the course of this work.

{
    \small
    \bibliographystyle{ieeenat_fullname}
    \bibliography{ref.bib}
}

\clearpage

\appendix

\section*{\Huge Appendix}

\section{Broader related work}

\paragraph{Self-supervised learning (SSL) of visual representations} has lately been of great interest to the scientific community, opening up the possibility of learning powerful models without labeled data~\citep{saleh2022selfsupervisedsurvey,balestriero2023cookbook}. SSL requires an appropriate \emph{pretext task} which replaces a data-defined objective, and over the years, a plethora of such tasks have been proposed~\cite{doersch2015unsupervised,zhang2016colorful,noroozi2017unsupervised,gidaris2018unsupervised}, with Joint-Embedding Architectures (JEA)~\cite{he2020momentum,chen2020simple,chen2020improved,caron2020unsupervised,zbontar2021barlow,grill2020bootstrap,chen2021exploring,chen2021empirical,caron2021emerging,zhou2022image,oquab2023dinov2}, and Masked image modeling (MIM)~\citep{vincent2008extracting,vincent2010stacked} gaining the most prominence in recent years. 

\paragraph{Limitations of JEA models} have been extensively covered by recent literature. JEA models rely on hand-crafted data augmentations~\cite{tian2020makes}, and their learned invariance to data perturbations can adversely affect the quality of representations~\cite{xiao2020whatshouldnotbecontrastive,lee2021improving,chavhan2023amortised,przewiezlikowski2024augmentation}. Moreover, JEA pretraining implicitly assumes a similar distribution of its pretraining and downstream task data~\cite{assran2023thehiddenuniform}, causing a need for additional dataset curation~\cite{oquab2023dinov2}. 
Therefore, development of SSL paradigms alternative to JEAs, including MIM, is an active line of research~\cite{assran2023ijepa,venkataramanan2024dora,garrido2024learningleveragingworldmodels,yuan2024semanticmim}.

\paragraph{Comparisons of Masked Image Modeling and Joint-Embedding Architectures} have been the focus of several works, which tried to understand the differences and combine the advantages of both paradigms~\cite{zhang2022howmask,kong2023understanding,park2023what,balestriero2024how}. The authors of \cite{zhang2022howmask,kong2023understanding} frame MIM as a JEA that learns invariance to image occlusions, but find its representation to be less expressive than in other JEAs. A theoretical study of learning by reconstruction, conducted in~\cite{balestriero2024how}, shows that data features required for reproducing pixels are misaligned with those needed for high-level perception. As a solution, multiple works propose shifting the prediction target from low-level pixels to higher-level image features, such as Histograms of Oriented Gradients~\cite{wei2023maskedfeaturepredictionselfsupervised} or latent representations~\cite{peng2022beitv2maskedimage,assran2023ijepa,liu2024exploring}, akin to the JEA objective.
Finally, \cite{park2023what} thoroughly compare the properties of MIM and JEA-trained models including, similarly to us, the attention mechanisms of their patch tokens. 
They find that whereas JEAs form global and homogeneous attention maps, the attention of MIM patch tokens is more localized. 
Furthermore, \cite{xie2023revealing,huang2024how} show that MIM-pretrained transformers produce attention patterns that capture diverse image aspects, useful for tasks which require spatial understanding of images.
Our work significantly extends these studies -- we analyze the \cls{} and patch representations of models trained with both paradigms and provide a detailed description of the information flow within them. 
We find that the attention mechanism emergent in MIM models imposes limitations that prevent these models from realizing their full potential in high-level perception tasks. 
Although this consequence of masked pretraining has previously been hinted at in the language models literature ~\cite{gao2021condenser}, to the best of our knowledge, it has not yet been discussed in the context of computer vision. While ~\cite{gao2021condenser} address this with a modified pretraining scheme, we present Selective Aggregation as a lightweight solution for improving existing MIM representations without requiring architectural changes or additional pretraining.

\section{Detailed experimental setup}

In this section, we describe our experimental methodology: our choice of pretrained models, the details and hyperparameters of evaluating their representations, as well as the codebase used for the experiments.

\subsection{Overview of the analyzed vision transformers}
\label{sec:official_vits}

Our study aims to verify whether Selective Aggregation of patch token representations with AbMILP can yield better representations than those of the \cls{} tokens.

For this purpose, we analyze various vision transformer architectures that were pretrained with several MIM and JEA approaches, using the parameters shared by the authors of the respective methods. This has two advantages:
\begin{itemize}
    \item Using the existing parameters significantly reduces the computational resources required for our study.
    \item Our study provides insights about the \emph{very same} sets of parameters that are described in their respective literature and used by the wider research community.
\end{itemize}
For a fair evaluation, we use the parameters of the models that were pretrained on the ImageNet-1k dataset~\cite{russakovsky2015imagenet}. All of the explored model parameters are compatible with the implementations of the MAE~\cite{he2021masked} or SimMIM~\cite{xie2022simmim} vision transformers.
Following the MAE and DINO implementations, when using ViT-S and ViT-B, we split the image into a $14\times14$ grid of patches of size $16\times16$. When using ViT-H, we split the image into $16\times16$ patches of size $14\times14$.

The 
only analyzed models that are not publicly available but were trained by us are the ViT-S pretrained with the MAE and the fine-tuned ViT-S/B/L variants of the MAE. To prepare these models, we used the MAE pretraining and fine-tuning codebase and hyperparameters~\cite{he2021masked}. 
Before fine-tuning, we initialize the model with the pretrained MAE parameters as shared by the authors and use the \cls{} token representation as input to the classifier.

\subsection{Representation evaluation details}
    \label{app:sec:hyperparams}
    In our evaluation of ViT representations in terms of classification accuracy on ImageNet-1k and other large datasets (NUS-WIDE, COCO, Food-101), we follow the MAE linear probing protocol~\cite{he2021masked}: we augment the images only by random cropping, use the batch size of 16,384, and train the classifier head for 90 epochs (50 in the case of ViT-Large and Huge) with the LARS optimizer~\cite{ginsburg2018large}, the base learning rate of 0.1 with cosine decay and 10 epochs of warmup, optimizer momentum of 0.9, and no weight decay. 
    For smaller datasets such as CUB, Stanford Cars, OxfordIIIPets, and ImageNet-1\%, we follow a similar linear probing setup but train using SGD with a batch size of 1024. We report the results averaged over 3 random seeds.
    When using the AbMILP Selective Aggregation, we train it alongside the classifier head.

    These evaluations are performed on a single node equipped with 4 NVIDIA-GH200 GPUs. Due to the memory constraints of this setup, we obtain the effective batch size of 16,384 by aggregating gradients from two forward passes with half of that batch size.

\subsection{Codebase}

    Our code is based on the official MAE codebase~\cite{he2021masked}, written in PyTorch~\cite{paszke2019pytorch}, and available at \href{https://github.com/gmum/beyond_cls}{\url{github.com/gmum/beyond_cls}}.
    We include scripts required for the analysis of the attention mechanism in ViTs, as well as linear evaluation of their representations extended with AbMILP~\cite{ilse2018attention}.

\section{Additional experimental results}

\subsection{Analysis of information flow in self-supervised ViT architectures}
    \label{sec:app:info_flow}
    This section contains the full details and experimental results of the attention mechanism in vision transformers, analyzed in \cref{sec:attn_analysis}. In the main manuscript, we include the analysis conducted on ViT-B, whereas in this section, we also provide the results of ViT-S and ViT-L architectures in \Cref{fig:cls_cls_attention_all_vits,fig:cls_pos_entropy_all_vits,fig:pos_self_attn_adj_all_vits,fig:pos_pos_entropy_all_vits}, For completeness, we re-include in them the pictograms describing each metric and the ViT-B results. We denote the contents of \Cref{fig:cls_cls_attention_all_vits,fig:cls_pos_entropy_all_vits,fig:pos_self_attn_adj_all_vits,fig:pos_pos_entropy_all_vits} in \cref{tab:attn_figures_corespondence}. Due to the size of the figures, include them at the end of this supplementary material.

To complement the analysis, we report the average entropy of attention from the \cls{} token to patch tokens across all Transformer blocks in~\cref{tab:cls_pos_entropy_agg}. These results extend our findings beyond MAE and confirm that MIM models tend to distribute \cls{} attention more uniformly across patch tokens compared to Joint-Embedding Architectures (JEAs). Notably, contemporary MIM architectures like I-JEPA and CAPI omit the \cls{} token altogether, using average pooling over patch tokens instead. This results in an effective attention distribution that is equivalent to uniform and thus exhibits entropy values near the theoretical upper bound.

    \begin{table}[h]

    \centering
              \resizebox{\linewidth}{!}{
    \begin{tabular}{cccccc|c}
        \toprule
        \textbf{Type} & \multicolumn{5}{c|}{\textbf{Masked Image Modeling}} & \textbf{JEA} \\
         \textbf{Model} & {MAE~\cite{he2021masked}} & {SimMIM~\cite{xie2022simmim}} & {BEIT-v2~\cite{peng2022beitv2maskedimage}} & {I-JEPA~\cite{assran2023ijepa}}$^\ddagger$ & {CAPI~\cite{darcet2025clusterpredictlatentspatches}}$^\ddagger$ & {DINO~\cite{caron2021emerging}} \\
         \midrule \midrule
         \textbf{Aggr. entropy} & \multirow{1}{*}{5.03} &  \multirow{1}{*}{4.96} &  \multirow{1}{*}{4.89} &  \multirow{1}{*}{5.28}  & \multirow{1}{*}{5.28} & \multirow{1}{*}{4.70} \\[-1ex]
          & & & & \multicolumn{2}{c|}{\shortstack{\small $\ddagger$ no \cls{} token -- uniform aggr.}} \\[-0.5ex]
         \bottomrule
    \end{tabular}
    }
    \caption{Patch aggregation entropy averaged across Transformer blocks in MIM models (we include DINO as a JEA representative for reference). MIMs aggregate patches more uniformly, motivating Selective Aggregation.}
    \label{tab:cls_pos_entropy_agg}
\end{table}

    \paragraph{Detailed methodology.} 

        In our analysis, we aim to characterize the attention patterns resulting from MIM and JEA pretraining. Therefore, for both \cls{} and patch tokens, we measure the attention of tokens to themselves (to see if tokens recycle their own information), and the entropy of attention to patch tokens (to see how information flows between the tokens).
        
        The entropy of an $i$-th token's attention to patch tokens (i.e. the $\mathbf{a}_{i,1:N}$ vector)
        is given by the Shannon entropy of its normalized values:
        \begin{align}
            \label{eq:entropy}
            \mathbb{H}(\mathbf{a}'_i) = -\sum\limits_{j=1}^N \mathbf{a}'_{i,j} \cdot log(\mathbf{a}'_{i,j}),
        \end{align}
        where $\mathbf{a}'_{i,1:N} = \frac{\mathbf{a}_{i,1:N}}{\sum\limits_{j=1}^N \mathbf{a}_{i,j}}$.
        We measure these values for each self-attention head in each ViT block and report the average results per block. The inference is performed on the ImageNet-1k validation dataset (50,000 images).

        To fairly compare Masked Image Modeling and Joint-Embedding paradigms, we analyze the ViT-B/16 models pretrained with MAE~\cite{he2021masked}, DINO~\cite{caron2021emerging}, MoCo-v3~\cite{chen2021empirical}, and iBOT~\cite{zhou2022image}, which represent prominent SSL approaches.\footnote{
        While iBOT optimizes a hybrid of JEA and MIM objectives, its performance gains are largely attributed to the JEA component~\cite{zhou2022image}, which is why we categorize it as such.}  
        We use publicly available pretrained parameters provided by their respective authors. 
        To examine whether optimizing for a global representation alters the attention behavior of MIM, we analyze an MAE model fine-tuned for ImageNet-1k classification using the \cls{} token.

    \paragraph{Analyzed models.} As discussed in \cref{sec:official_vits}, whenever possible, for each analyzed method, we use the ImageNet-1k pretrained model parameters officially released by their respective authors. The only exception to this is the MAE trained with ViT-S, which we trained ourselves, and the finetuned MAE (MAE-FT), which we finetuned ourselves for ImageNet-1k classification on top of the \cls{} token features. 
    Due to the lack of available ViT-L parameters of MoCo-v3~\cite{chen2021empirical} and DINO~\cite{caron2021emerging}, we omit them from the analysis of this architecture. 
    However, given that the three JEA approaches behave similarly for each property analyzed in ViT-S and ViT-B architectures, we believe that the available ViT-L iBOT~\cite{zhou2022image} variant sufficiently represents JEA. 
    Similarly, we do not conduct this comparison with the ViT-H architecture, due to the lack of publicly available parameters of ViT-H trained with JEA to compare with.

    \paragraph{Discussion.} 

    We are interested in the behavior of the ViT attention mechanism emergent in the MAE and JEA approaches, especially in the deep ViT blocks which form higher-level image representations~\cite{yosinski2014transferable}. Across the three ViT architectures analyzed, we observe several consistent trends, more generally discussed in \Cref{sec:attn_analysis} and summarized below:
    \begin{itemize}
        \item The \cls{} token of pretrained and fine-tuned MAE assigns a large portion of attention (around 40-50\%) to its own representation.
        \item The entropy of attention between the \cls{} and patch tokens is much higher in MAE than in the rest of the models, indicating that it aggregates the information from a larger number of patch tokens. 
        Fine-tuning of the MAE decreases this value to the levels observed in JEA models, increasing the selectiveness of attention.
        \item The attention of MAE patch tokens to themselves (relative to all patch tokens) is higher than in other models, indicating they are more likely to preserve their own, diverse information~\cite{park2023what}. Fine-tuning of the MAE results in lowering this metric to the level observed in the JEA models. MAE patches also attend to other patches with lower entropy than in JEAs and this does not change after fine-tuning.
    \end{itemize}

    \begin{table}[]
        \centering
        \resizebox{\linewidth}{!}{
            \begin{tabular}{ccc}
            \toprule
            \multirow{2}{*}{\textbf{Metric}} & \textbf{ViT-B results} & \textbf{ViT-S/B/L results} \\
            & \textbf{(manuscript)} & \textbf{(Appendix)} \\
            \midrule \midrule
            \cls{}-\cls{} attention & \cref{fig:cls_cls_attention} & \cref{fig:cls_cls_attention_all_vits} \\
            \cls{}-patch entropy &  \cref{fig:cls_pos_entropy} & \cref{fig:cls_pos_entropy_all_vits} \\
            patch-patch attention & \cref{fig:pos_self_attn_adj} & \cref{fig:pos_self_attn_adj_all_vits} \\
            patch-patch entropy & \cref{fig:pos_pos_entropy} & \cref{fig:pos_pos_entropy_all_vits} \\
            \bottomrule
            \end{tabular}
        }
        \caption{A reference of Figures depicting the analysis of the attention mechanism and their extended counterparts in the Appendix.}
        \label{tab:attn_figures_corespondence}
    \end{table}

\subsection{Designing the token aggregation mechanism}
\label{sec:ablation}

In this section, we discuss different design choices for the token aggregation function, which uses either various variants of AbMILP~\cite{ilse2018attention}, or other, non-trainable substitutes.
Unless specified otherwise, all experiments reported in this section are conducted with the ViT-B model pretrained with the MAE~\cite{he2021masked}.

\paragraph{Ablation study of AbMILP variants.}
We explore several designs of the model used by AbMILP to predict the scores for patch aggregation and report their performance in~\cref{tab:abmilp_ablation}.
\begin{table}[h]
    \centering
    \begin{tabular}{ccccc}
        \toprule
        \textbf{Activation} & \multicolumn{4}{c}{\textbf{AbMILP MLP depth}} \\
        \textbf{function} & \textbf{1}$^\dagger$ & \textbf{2} & \textbf{3} & \textbf{4} \\
        \midrule \midrule
         ReLU &  \multirow{3}{*}{\shortstack{71.6\\ \footnotesize $^\dagger$linear model \\ \footnotesize w/o activation}} & 71.7 & 71.5 & 71.6 \\
         GeLU &  & 71.6 & 71.6 & 71.5 \\
         Tanh &  & 68.7 & 66.7 & 66.7 \\
         \bottomrule
    \end{tabular}
    \caption{Comparison of ImageNet-1k classification accuracy of the MAE representation aggregated by different variants of AbMILP~\cite{ilse2018attention}. Deeper MLPs do not boost performance.}
    \label{tab:abmilp_ablation}
\end{table}

The original AbMILP architecture~\cite{ilse2018attention} uses a 2-layer MLP with the Tanh activation function. MAE patch tokens aggregated by this model achieve an accuracy of 68.7\%. Although this is higher than the \cls{} token representation, we found that the training process is unstable and replaced the Tanh activation with ReLU. This led to more stable training and an improvement in accuracy by almost 3 pp. Surprisingly, reducing the MLP to a single linear layer achieves almost the same results. 
Due to the simplicity and performance of this design, we adopt it in our main experiments. As seen in \cref{sec:main_experiments}, 
the effectiveness of this approach generalizes to 
aggregating representations of MIM models other than the MAE.

We note that AbMILP is just one of several Multiple-Instance Learning methods that can be adopted to aggregate patch token representations. As an alternative, we explore the Self-Attention AbMILP~\cite{rymarczyk2021kernel} where, prior to computing the aggregation scores and the aggregated representation, tokens are processed by an additional trainable self-attention head.
This approach achieves accuracy much closer to that of the JEA-trained approaches -- 74.83\%.
This indicates an even larger richness of information stored in the representation space of Masked models, which requires more complex task-specific heads in order to be fully exploited. However, we found the training of this model to be unstable with the LARS optimizer~\cite{ginsburg2018large}, and were only able to train it using SGD.
Moreover, a classification head that internally uses trainable self-attention to pre-process the classifier input is incomparable to a simple linear probe. For these reasons, we do not include this approach in our main experiments.

\newcommand\avgmae{Average MAE \cls{} token attention}
\newcommand\leastentropymae{Lowest-entropy MAE \cls{} token attention}
\newcommand\centralpatch{MAE central patch token attention}
\newcommand\avgdino{Average DINO \cls{} token attention}
\newcommand\abmilp{AbMILP}

\begin{figure*}[h]
    \centering
    {
        \footnotesize
        \parbox{0.15\linewidth}{\centering Input image} 
        \hfill 
        \parbox{0.15\linewidth}{\centering \avgmae}
        \hfill 
        \parbox{0.15\linewidth}{\centering \leastentropymae}
        \hfill 
        \parbox{0.15\linewidth}{\centering \centralpatch}
        \hfill 
        \parbox{0.15\linewidth}{\centering \avgdino}
        \hfill 
        \parbox{0.15\linewidth}{\centering \abmilp}
    }

    \includegraphics[width=\linewidth]{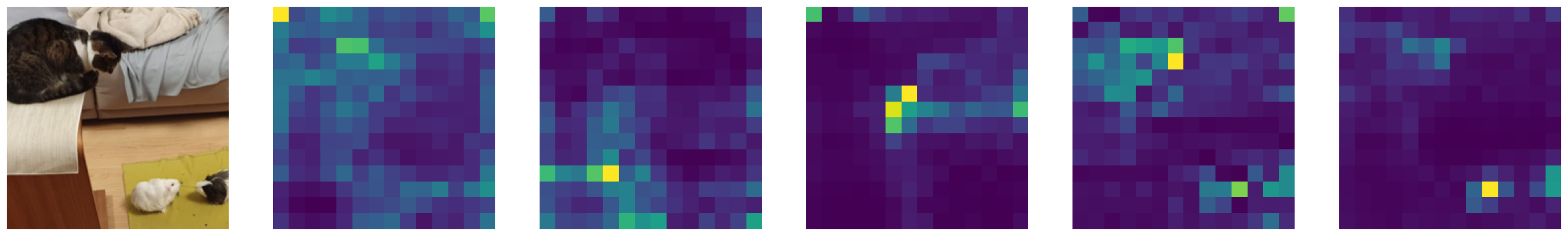}
    \includegraphics[width=\linewidth]{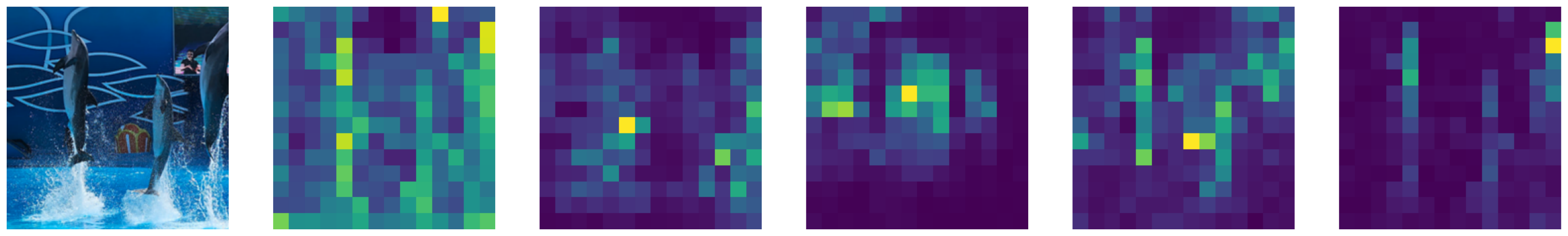}
    \includegraphics[width=\linewidth]{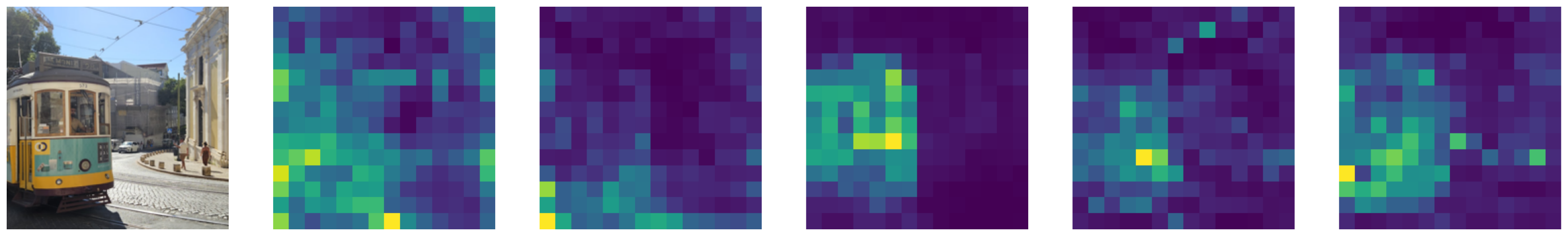}
    \includegraphics[width=\linewidth]{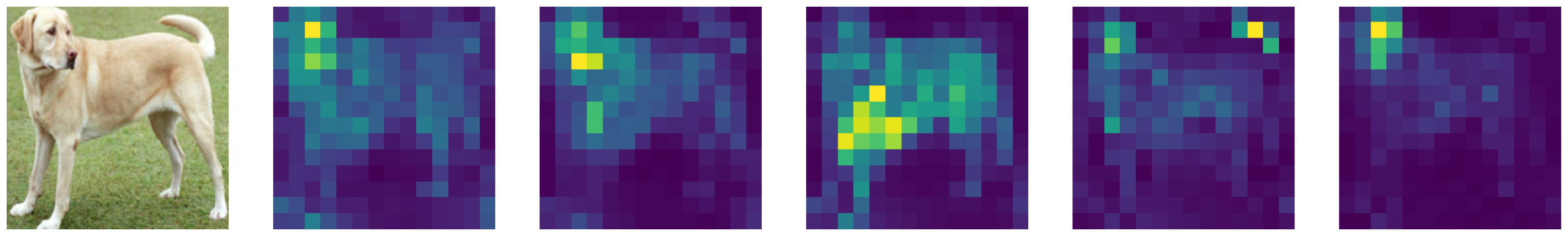}
    \includegraphics[width=\linewidth]{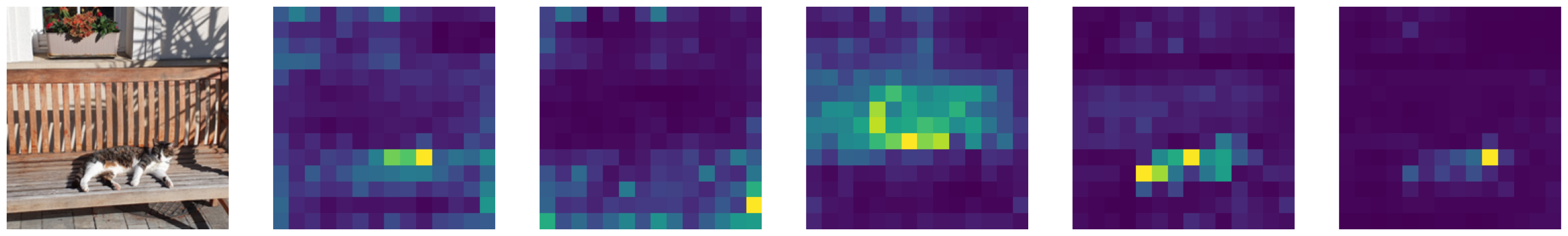}
    \includegraphics[width=\linewidth]{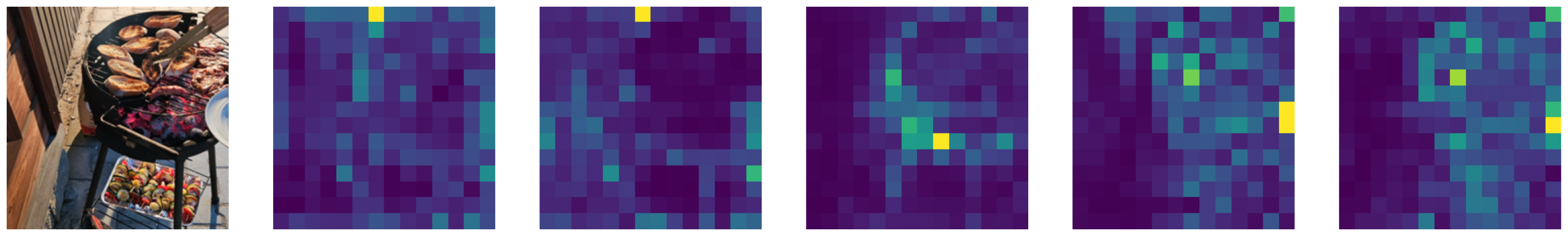}

    \caption{Example token aggregation scores produced by different approaches denoted in columns. The average \cls{} attention of the MAE aggregates the patches too uniformly. The \cls{} attention with lowest entropy and the attention of the central patch have low entropy, but are not guaranteed to capture the object of interest in the image. Finally, the DINO \cls{} attention maps and aggregation vectors produced by AbMILP reliably identify the most crucial patches for forming high-level global image representations.
    }
    \label{fig:token_selector_attentions}
\end{figure*}

\paragraph{Non-trainable token aggregation.}
Apart from the AbMILP-based aggregation, we explore several alternative token aggregation functions that are not trained along with the classifier model. We discuss these approaches and their properties below and report their representations' average accuracies and entropies of the aggregation vectors  in~\cref{tab:non_trainable_token_selection}. To measure if different token aggregation approaches select the same patch tokens, in \cref{fig:attn_kld}, we report the average Kullback-Leibler Divergence between token selection vectors produced by each method. 
Finally, we visualize the example token selection vectors in~\cref{fig:token_selector_attentions}.

\begin{itemize}
    \item \textbf{\avgmae} -- the average attention between the \cls{} and patch tokens, produced by the MSA of the final MAE ViT block. As evidenced by the high entropy, this approach aggregates many patches, achieving quality similar to that of the regular \cls{} representation.
    \item \textbf{\leastentropymae} -- the attention map between the \cls{} and patch tokens produced by the MSA of the final MAE ViT block, which has the lowest entropy. This approach achieves low aggregation entropy, but due to the diversity of image fragments attended by different self-attention heads~\cite{park2023what}, the attended fragment of an image is not guaranteed to contain the object of interest.
    \item \textbf{\centralpatch} -- the average attention between the token of the central patch in the image and other patches. This approach can distinguish the tokens of the object of interest as long as it is depicted on the central image patch, which is not always the case.
    As evidenced by the high KLD between the \leastentropymae ~and \centralpatch, these two approaches tend to have a low agreement in terms of which tokens to select, suggesting their high volatility.
    \item \textbf{\avgdino}  -- the average attention between the \cls{} and patch tokens, produced by the MSA of the final DINO ViT block. As observed by~\cite{caron2021emerging}, DINO attention maps are exceptionally good at capturing the main objects of interest in the images. MAE patch tokens selected with this approach form representations superior to the \cls{} token, but an obvious drawback of this approach is the reliance on an externally pretrained model. As seen in ~\cref{fig:attn_kld}, this selects tokens most similar to the AbMILP-based token aggregation.
\end{itemize}

\begin{table}[h]
        \centering
        \resizebox{\linewidth}{!}{
        \begin{tabular}{ccc}
            \toprule
            \textbf{Token aggregation approach} & \textbf{Accuracy} & \textbf{Entropy} \\
            \midrule \midrule
            \avgmae & 67.8 & 5.14 \\
            \leastentropymae & 66.3 & 4.77 \\
            \centralpatch  & 65.2 & 4.70 \\
            \avgdino & 70.9 & 4.89 \\
            \abmilp & 71.6 & 4.80 \\
             \bottomrule
        \end{tabular}
        }
        \captionof{table}{
        Evaluation of different token aggregation approaches in terms of classification accuracy of their representations, and entropy of the aggregation vectors they produce.
        }
        \label{tab:non_trainable_token_selection}
\end{table}
\begin{figure}[h]
        \centering
        \resizebox{\linewidth}{!}{
        \begin{tikzpicture}
    \begin{axis}[
        width=0.75\linewidth, 
        height=0.75\linewidth, 
        colormap/hot2,              %
        colorbar,                      %
        colorbar style={
            yticklabel style={/pgf/number format/fixed},
        },
        xtick={0,1,2,3,4},             %
        ytick={0,1,2,3,4},             %
        xticklabels={
            \textbf{a)} , 
            \textbf{b)}, 
            \textbf{c)}, 
            \textbf{d)}, 
            \textbf{e)}
        },   %
        yticklabels={
            \textbf{a)} \parbox{0.4\linewidth}{\raggedleft \avgmae}, 
            \textbf{b)} \parbox{0.4\linewidth}{\raggedleft \leastentropymae}, 
            \textbf{c)} \parbox{0.4\linewidth}{\raggedleft \centralpatch},
            \textbf{d}) \parbox{0.4\linewidth}{\raggedleft \avgdino},
            \textbf{e)} \parbox{0.4\linewidth}{\raggedleft \abmilp}
        },   %
        y dir=reverse,
        xmin=-0.5, xmax=4.5,
        ymin=-0.5, ymax=4.5,
        enlarge x limits=false,
        enlarge y limits=false,
        point meta=explicit,  %
        point meta min=0, point meta max=0.85,  %
        nodes near coords,                %
        nodes near coords align={center},  %
        every node near coord/.append style={
            text=black                      %
        },
        colormap={reverse hot2}{
            indices of colormap={
                \pgfplotscolormaplastindexof{hot2},...,0 of hot2}
        },
    ]

    \addplot [matrix plot*, mesh/cols=5, mesh/rows=5] table [meta expr=\thisrow{val}] {
        x y val
        0 0 0.0
        1 0 0.24836694
        2 0 0.44806281
        3 0 0.23861243
        4 0 0.24597117
        0 1 0.23060021
        1 1 0.0
        2 1 0.71483876
        3 1 0.47609598
        4 1 0.44809859
        0 2 0.42086509
        1 2 0.69382341
        2 2 0.0
        3 2 0.4507144
        4 2 0.46542974
        0 3 0.23012565
        1 3 0.47950203
        2 3 0.49668083
        3 3 0.0
        4 3 0.23929513
        0 4 0.24284151
        1 4 0.44202676
        2 4 0.4899629
        3 4 0.23643296
        4 4 0.0
    };
    \end{axis}
\end{tikzpicture}
        }
        \caption{Mean KLD between aggregation vectors produced by different token aggregation techniques.}
        \label{fig:attn_kld}
\end{figure}

Most of the above approaches select the MAE patch tokens with an entropy close to that observed in the JEA \cls{} token. 
However, except for the attention maps generated by DINO and AbMILP, we did not find an approach that would reliably select patch tokens to form a representation of better quality than the \cls{} token. Finding such tokens in an unsupervised manner is an interesting direction for future work.

\paragraph{Selective Aggregation and Attentive Probing}

Attentive Probing (AP)~\cite{chen2023context} has been proposed as an alternative to naive feature aggregation in ViTs. Similarly to our Selective Aggregation, AP learns to emphasize the most relevant patch tokens while keeping the encoder parameters frozen. However, AP differs from our approach in a key way: it does not only learn to aggregate tokens, but also transforms them with a cross-attention layer into a new representation space. potentially more suitable for the downstream task~\cite{balestriero2024how}. In contrast, AbMILP is designed to isolate the aggregation process while preserving the original ViT representations.

We compare AP and AbMILP across multiple MIM models in terms of ImageNet-1k classification and report the results in \cref{tab:attn_probing}. Since AP typically uses a 12-head self-attention mechanism, we additionally evaluate a reduced variant with a single attention head (without reducing the representation dimensionality) to better compare with the capacity of AbMILP (which predicts a single set of representation aggregation weights). As expected, the full AP model achieves the best results, benefiting from its greater expressive power. However, despite AP's significantly higher parameter and compute cost, reducing it to a single head brings its performance in line with AbMILP. This result is somewhat surprising and suggests that AP’s strength may come from ensembling multiple Selective Aggregation patterns rather than from the learned transformation. Exploring this insight to develop more efficient Selective Aggregation strategies is a promising direction for future work.

\begin{table}[h]
    \centering
      \resizebox{\linewidth}{!}{
    \begin{tabular}{ccccc}
        \toprule
        \multicolumn{2}{c}{\textbf{Encoder}} & \multicolumn{3}{c}{\textbf{Aggregation method}} \\
        {Initialization} & ViT type &  AbMILP & AP (1 head) & AP (12 heads) \\ %
         \midrule \midrule 
        MAE~\cite{he2021masked} & ViT-S &  {54.4} & 53.6 & \textbf{63.9} \\
        MAE~\cite{he2021masked} & ViT-B &  {71.6} & 71.4& \textbf{75.4} \\
        MAE~\cite{he2021masked} & ViT-L &  {77.4} & 77.6 & \textbf{79.7} \\
        MAE~\cite{he2021masked} & ViT-H &  {78.1} & 78.3 & \textbf{80.0} \\

        BEIT-v2~\cite{peng2022beitv2maskedimage} & ViT-B & {80.9} & 81.0 & \textbf{81.8}\\
        I-JEPA~\cite{assran2023ijepa} & ViT-H &  {79.2} & 79.5 & \textbf{79.7} \\
        CAPI~\cite{darcet2025clusterpredictlatentspatches} & ViT-L &  82.4 & 81.6 & \textbf{82.7}\\
        \midrule
        \bottomrule
    \end{tabular}
    }
    \caption{Comparison of AbMILP~\cite{ilse2018attention} and Attentive Probing (AP)~\cite{chen2023context} aggregation schemes. AbMILP and the single-head cross-attention AP perform comparably.}
    \label{tab:attn_probing}
\end{table}

\subsection{Using Selective Aggregation for object localization.} 
\label{app:sec:localization}
\begin{figure*}[t]
    \begin{minipage}{0.33\textwidth}
        \includegraphics[width=\textwidth, height=4cm]{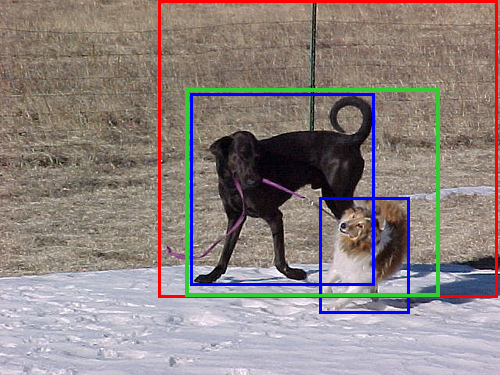}
    \end{minipage}
    \hfill
    \begin{minipage}{0.33\textwidth}
        \includegraphics[width=\textwidth, height=4cm]{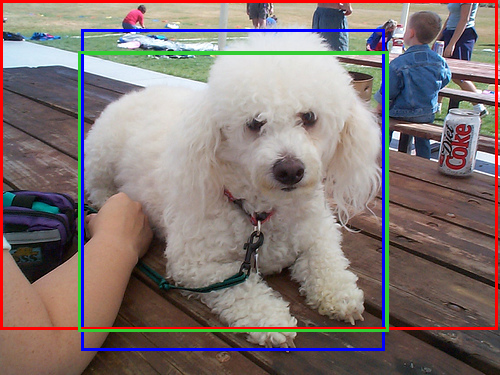}
    \end{minipage}
    \hfill
    \begin{minipage}{0.33\textwidth}
        \includegraphics[width=\textwidth, height=4cm]{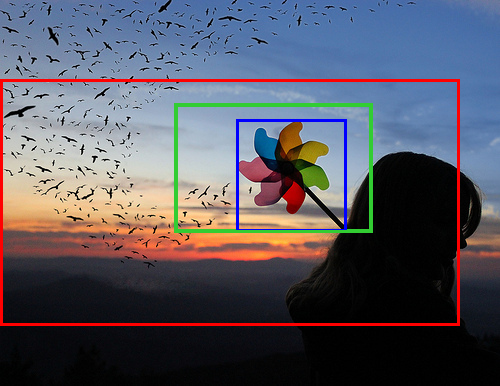}
    \end{minipage}
    \begin{minipage}{0.33\textwidth}
        \includegraphics[width=\textwidth, height=4cm]{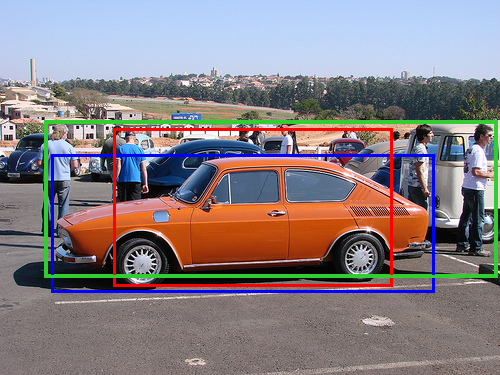}
    \end{minipage}
    \hfill
    \begin{minipage}{0.33\textwidth}
        \includegraphics[width=\textwidth, height=4cm]{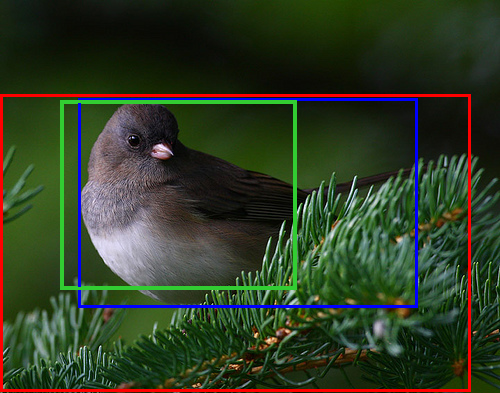}
    \end{minipage}
    \hfill
    \begin{minipage}{0.33\textwidth}
        \includegraphics[width=\textwidth, height=4cm]{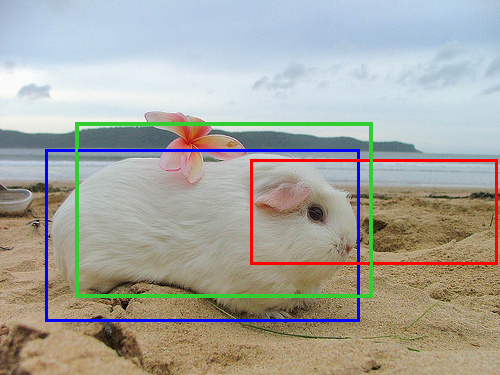}
    \end{minipage}
    \caption{Example localization results of the MAE \cls{} attention and Selective Aggregation weights. {\color{blue} Blue}: ground-truth. {\color{red} Red}: bounding box predicted from the \cls{} attention map. {\color{green} Green}: bounding box predicted from the Selective Aggregation scores. Selective Aggregation locates objects with better accuracy (see~\cref{tab:localization}).}
    \label{fig:localization}
\end{figure*}

While global representations, which are the focus of this paper, are not generally suitable for dense prediction tasks, their attention maps can be used as a means to localize the object of interest in the image~\cite{caron2021emerging}. Because Selective Aggregation highlights the most relevant tokens, it can be used in a similar manner. We evaluate this capability of Selective Aggregation with the MAE and BEIT-v2 models, comparing it to their \cls{} attention maps. We measure the localization quality in terms of MaxBoxAccV2~\cite{choe2020evaluating,psomas2023simpool} on the ImageNet validation dataset. We report the results in \cref{tab:localization}, and visualize the example results in~\cref{fig:localization}. Our results indicate that the more focused Selective Aggregation localizes the objects of interest more accurately. 

\begin{table}[t]
    \centering
      \resizebox{\linewidth}{!}{
    \begin{tabular}{cccc}
        \toprule
       \multicolumn{2}{c}{\textbf{Encoder}} & \multicolumn{2}{c}{\textbf{Localization based on}} \\
       {Source} & {ViT} & \cls{} attention map & Selective Aggregation map \\ %
         \midrule \midrule
        MAE~\cite{he2021masked} & ViT-B & 53.3 & \textbf{59.4}  \\
        BEIT-v2~\cite{peng2022beitv2maskedimage} & ViT-B & 44.3 & \textbf{65.1} \\
        \bottomrule
    \end{tabular}
    }
    \caption{Object localization capabilities of the \cls{} attention and Selective Aggregation weights, measured in terms of MaxBoxAccV2~\cite{choe2020evaluating} on the ImageNet validation dataset.}  
    \label{tab:localization}
\end{table}

\section{Future research directions}

Our results indicate that lack of global representation aggregation is inherent to vision transformers trained with Masked Image Modeling. In this section, we summarize several potential research directions for better understanding this issue.

\paragraph{Unsupervised discovery of relevant tokens.}
We have showed that a shallow AbMILP~\cite{ilse2018attention} is sufficient for recognizing the patch tokens of MIM models that are relevant to form global image representations. However, in each MIM model, we learn that function together with the classifier dedicated to downstream tasks. Understanding what makes a patch token relevant for global representation and finding such tokens in an unsupervised manner is a natural further direction.

\paragraph{Scaling Selective Aggregation.} Our implementation uses the minimal version of the aggregation score prediction model. 
In our comparison with Attentive Probing, we show that it succeeds not necessarily due to further processing of representations, but rather due to an ensemble of multiple self-attention heads. A full study of the effectiveness of vertical (more complex transformations) and horizontal (larger ensemble of aggregation functions) scaling of Selective Aggregation would be very beneficial for determining the most efficient MIM adaptation protocol.

\paragraph{Aggregation of internal ViT representations.}
Currently, Selective Aggregation acts only on patch representations of the final ViT block. 
While this approach improves the MIM representations, we note that it does not interfere in any way with their internal information flow.  However, as shown in \cref{fig:cls_pos_entropy}, the \cls{} token of JEAs aggregates patch information increasingly selectively throughout the several final model blocks. 
We hypothesize that similarly aggregating MIM representations within internal ViT blocks, either via additional training objectives or post-pretraining modifications, could yield further improvements in their quality.

    \pgfplotsset{ylabel/.style={}}

    \begin{figure*}
    \centering
    \begin{minipage}{0.48\linewidth}
        \centering
        \begin{tikzpicture}[
    node distance=1.25cm and 0.75cm,
]

    \node[draw] (zin0) {$\mathbf{v}_0$};
    \node[draw, right=1cm of zin0, gray] (zin1) {$\mathbf{v}_1$};
    \node[right=0.25cm of zin1, gray] (zindots) {$\dots$};
    \node[draw, right=0.25cm of zindots, gray] (zinN) {$\mathbf{v}_N$};

    \node[draw, below=1.7cm of zin0] (zout0) {$\mathbf{o}_0$};
    \node[right=of zout0] (zoutdots) {$\dots$};

    \draw[-stealth, thick] (zin0) -- (zout0) node[midway, fill=white, inner sep=3pt, draw=red, dashed, very thick, rounded corners] (a00) {$\mathbf{a}_{0,0}$};
    
    \draw[-stealth, thick, gray] (zin1.south) -- (zout0) node[midway, fill=white, inner sep=2pt] {$\mathbf{a}_{0,1}$};
    \draw[-stealth, thick, gray] (zinN.south) -- (zout0) node[midway, fill=white, inner sep=2pt] {$\mathbf{a}_{0,N}$};

    \draw[decorate,decoration={brace,amplitude=6pt}] 
        ([yshift=5pt]zin0.north west) -- ([yshift=5pt]zin0.north east) 
        node[midway, above=6pt] {\cls{} token};

    \draw[decorate,decoration={brace,amplitude=6pt},gray] 
        ([yshift=5pt]zin1.north west) -- ([yshift=5pt]zinN.north east) 
        node[midway, above=6pt] {Patch tokens};

    \node[left=of zin0, align=center] (int) {Value\\tokens};
    \node[below=of int, align=center] (outt) {Output\\tokens};
    \draw[-stealth, thick] (int) -- (outt) node[midway, fill=white, inner sep=1pt,align=center] {\scriptsize Attention};

\end{tikzpicture}
    \end{minipage}
    \hfill
    \begin{minipage}{0.48\linewidth}
        \centering
            \begin{tikzpicture}
        \begin{axis}[
            sec4style,
            ylabel={{\shortstack{Attention of [cls] token to itself}}},
	xlabel={ViT-S block}
	]
 \addplot[{DINOStyle}]
 table {%
1 0.22592461109161377
2 0.4610525667667389
3 0.8372762203216553
4 0.5499194860458374
5 0.5324665904045105
6 0.2569330334663391
7 0.18351706862449646
8 0.18615882098674774
9 0.15426552295684814
10 0.15333449840545654
11 0.11278890818357468
12 0.09421324729919434
};
\addlegendentry{DINO~\cite{caron2021emerging}}

 \addplot[{iBOTStyle}]
 table {%
1 0.14220421016216278
2 0.6167657375335693
3 0.8530445098876953
4 0.5024808645248413
5 0.5201621651649475
6 0.3940814137458801
7 0.24969255924224854
8 0.22873570024967194
9 0.13550913333892822
10 0.121304452419281
11 0.07956995069980621
12 0.07573717087507248
};
\addlegendentry{iBOT~\cite{zhou2022image}}

 \addplot[{MoCoStyle}]
 table {%
1 0.6865448951721191
2 0.10935171693563461
3 0.6983090043067932
4 0.5431168675422668
5 0.36661553382873535
6 0.41509729623794556
7 0.272640585899353
8 0.4074997901916504
9 0.48094624280929565
10 0.21007759869098663
11 0.04595419019460678
12 0.015424410812556744
};
\addlegendentry{MoCo-v3~\cite{chen2021empirical}}

 \addplot[{MAEStyle}]
 table {%
1 0.5003588199615479
2 0.5123068690299988
3 0.4841952621936798
4 0.5394871234893799
5 0.5982891321182251
6 0.5502391457557678
7 0.5316312909126282
8 0.33505362272262573
9 0.4515310823917389
10 0.2836794853210449
11 0.42536044120788574
12 0.343818724155426
};
\addlegendentry{MAE~\cite{he2021masked}}

 \addplot[{FinetunedMAEStyle}]
 table {%
1 0.6730453968048096
2 0.6564021110534668
3 0.6233834624290466
4 0.5832746624946594
5 0.42043083906173706
6 0.3456310033798218
7 0.43262770771980286
8 0.3689168095588684
9 0.4072069823741913
10 0.3567304313182831
11 0.28023749589920044
12 0.3613455593585968
};
\addlegendentry{MAE (FT)~\cite{he2021masked}}

\end{axis}
\end{tikzpicture}
    \end{minipage}
    \\
    \begin{minipage}{0.48\linewidth}
            \begin{tikzpicture}
        \begin{axis}[
            sec4style,
            ylabel={{\shortstack{Attention of the\\\cls{}token to itself}}},
	xlabel={ViT-B block}
	]
 \addplot[{DINOStyle}]
 table {%
1 0.1585644781589508
2 0.4371647536754608
3 0.4487747251987457
4 0.2436964064836502
5 0.08758749067783356
6 0.10344899445772171
7 0.1381075233221054
8 0.1890965849161148
9 0.18715742230415344
10 0.1650734394788742
11 0.1659216284751892
12 0.18133679032325745
};
\addlegendentry{DINO~\cite{caron2021emerging}}

 \addplot[{iBOTStyle}]
 table {%
1 0.17956870794296265
2 0.42291003465652466
3 0.6281848549842834
4 0.3125123083591461
5 0.14374788105487823
6 0.1439746916294098
7 0.10955562442541122
8 0.10184510052204132
9 0.1253819465637207
10 0.11884170770645142
11 0.11997197568416595
12 0.11452257633209229
};
\addlegendentry{iBOT~\cite{zhou2022image}}

 \addplot[{MoCoStyle}]
 table {%
1 0.3921937048435211
2 0.37001293897628784
3 0.5669093132019043
4 0.27546176314353943
5 0.19073139131069183
6 0.10868073999881744
7 0.11473498493432999
8 0.1351258009672165
9 0.09406556189060211
10 0.09086806327104568
11 0.05742160230875015
12 0.08490914106369019
};
\addlegendentry{MoCo-v3~\cite{chen2021empirical}}

\addplot[{MAEStyle}]
 table {%
1 0.4105907082557678
2 0.5290440917015076
3 0.40943580865859985
4 0.5888752341270447
5 0.540928304195404
6 0.7101895213127136
7 0.5992339849472046
8 0.6591408252716064
9 0.479407399892807
10 0.4309639036655426
11 0.5721790790557861
12 0.4142257571220398
};
\addlegendentry{MAE~\cite{he2021masked}}

 \addplot[{FinetunedMAEStyle}]
 table {%
1 0.6157293915748596
2 0.6317875981330872
3 0.6273791193962097
4 0.719139814376831
5 0.5765323042869568
6 0.6416050791740417
7 0.7379360795021057
8 0.7787177562713623
9 0.7339198589324951
10 0.710572361946106
11 0.6700039505958557
12 0.6385836005210876
};
\addlegendentry{MAE (FT)~\cite{he2021masked}}

\end{axis}
\end{tikzpicture}
    \end{minipage}
    \hfill
    \begin{minipage}{0.48\linewidth}
            \begin{tikzpicture}
        \begin{axis}[
            sec4style,
            ylabel={{\shortstack{Attention of the\\\cls{}token to itself}}},
	xlabel={ViT-L block}
	]
 \addplot[{iBOTStyle}]
 table {%
1 0.24908284842967987
2 0.17164337635040283
3 0.3152023255825043
4 0.39708730578422546
5 0.17031528055667877
6 0.22046220302581787
7 0.10381939262151718
8 0.1633441150188446
9 0.1361808329820633
10 0.0959344357252121
11 0.11675063520669937
12 0.1615898758172989
13 0.17874425649642944
14 0.18223132193088531
15 0.18679893016815186
16 0.19030611217021942
17 0.1783057004213333
18 0.17706847190856934
19 0.17781409621238708
20 0.19622047245502472
21 0.22782345116138458
22 0.21613167226314545
23 0.16828159987926483
24 0.15034839510917664
};
\addlegendentry{iBOT~\cite{zhou2022image}}

\addplot[{MAEStyle}]
 table {%
1 0.4678254723548889
2 0.5214612483978271
3 0.5682708024978638
4 0.5421594977378845
5 0.4499031901359558
6 0.38204509019851685
7 0.526643693447113
8 0.5374982953071594
9 0.525947093963623
10 0.759601891040802
11 0.6261979341506958
12 0.5756962895393372
13 0.6277931928634644
14 0.5179585814476013
15 0.559093177318573
16 0.4985325336456299
17 0.5067914724349976
18 0.5401639938354492
19 0.5563744306564331
20 0.5677240490913391
21 0.49095281958580017
22 0.51912522315979
23 0.5493794679641724
24 0.487460196018219
};
\addlegendentry{MAE~\cite{he2021masked}}

 \addplot[{FinetunedMAEStyle}]
 table {%
1 0.6022951006889343
2 0.6604572534561157
3 0.6791107654571533
4 0.6422051191329956
5 0.6186037063598633
6 0.5138368606567383
7 0.6696961522102356
8 0.6519820094108582
9 0.5790942311286926
10 0.8023564219474792
11 0.7331181764602661
12 0.6688432097434998
13 0.8032627105712891
14 0.7791123986244202
15 0.8573797941207886
16 0.8815737366676331
17 0.8822301626205444
18 0.8792131543159485
19 0.8744639158248901
20 0.8795086741447449
21 0.8645806908607483
22 0.8587735295295715
23 0.8526384234428406
24 0.8090367317199707
};
\addlegendentry{MAE (FT)~\cite{he2021masked}}

\end{axis}
\end{tikzpicture}
    \end{minipage}
    \caption{
    Extended version of \Cref{fig:cls_cls_attention}.
    Attention of the \cls{} token to itself is much higher in both pretrained and finetuned MAE, than in the JEA ViTs. As opposed to JEA, where the \cls{} tokens gather a large amount of information from the patch tokens, the MAE \cls{} token primarily recycles its own representation.
    }
    \label{fig:cls_cls_attention_all_vits}
\end{figure*}

\begin{figure*}
    \centering
    \begin{minipage}{0.48\linewidth}
        \centering
        \begin{tikzpicture}[
    node distance=1.25cm and 0.75cm,
]

    \node[draw, gray] (zin0) {$\mathbf{v}_0$};
    \node[draw, right=1 cm of zin0] (zin1) {$\mathbf{v}_1$};
    \node[right=of zin1] (zindots) {$\dots$};
    \node[draw, right=of zindots] (zinN) {$\mathbf{v}_N$};

    \node[draw, below=1.7 cm of zin0] (zout0) {$\mathbf{o}_0$};
    \node[right=of zout0] (zoutdots) {$\dots$};

    \draw[-stealth, thick, gray] (zin0) -- (zout0) node[midway, fill=white, inner sep=0pt] {$\mathbf{a}_{0,0}$};
    \draw[-stealth, thick] (zin1) -- (zout0) node[midway, fill=white, inner sep=2pt] (a01){$\mathbf{a}_{0,1}$};
    \draw[-stealth, thick] (zinN) -- (zout0) node[midway, fill=white, inner sep=2pt] (a0N) {$\mathbf{a}_{0,N}$};
    \node[left=0.01cm of a0N] {$\dots$};

    \draw[decorate,decoration={brace,amplitude=6pt}, gray] 
        ([yshift=5pt]zin0.north west) -- ([yshift=5pt]zin0.north east) 
        node[midway, above=6pt] {\cls{} token};

    \draw[decorate,decoration={brace,amplitude=6pt}] 
        ([yshift=5pt]zin1.north west) -- ([yshift=5pt]zinN.north east) 
        node[midway, above=6pt] {Patch tokens};

    \node[left=of zin0, align=center] (int) {Value\\tokens};
    \node[below=of int, align=center] (outt) {Output\\tokens};
    \draw[-stealth, thick] (int) -- (outt) node[midway, fill=white, inner sep=1pt,align=center] {\scriptsize Attention};

    \node[draw=red, dashed,fit=(a01) (a0N), very thick, inner sep=1pt, rounded corners] (a_container) {};

    \node[below right=0.1cm and 0.5cm of a_container, red] (ent_label) {Entropy}; %
    
    \draw[-stealth, very thick, dashed, red] (a_container) -- (ent_label); %
    
\end{tikzpicture}
    \end{minipage}
    \hfill
    \begin{minipage}{0.48\linewidth}
        \centering
            \begin{tikzpicture}
        \begin{axis}[
            sec4style,
            ylabel={{\shortstack{Entropy of normalized attention
between [cls] and patch tokens}}},
	xlabel={ViT-S block}
	]
 \addplot[{DINOStyle}]
 table {%
1 5.108044624328613
2 5.09736442565918
3 4.85870361328125
4 5.060513019561768
5 5.0546112060546875
6 5.123851776123047
7 5.016471862792969
8 4.973426342010498
9 4.8105316162109375
10 4.659380912780762
11 4.655857086181641
12 4.771890640258789
};
\addlegendentry{DINO~\cite{caron2021emerging}}

 \addplot[{iBOTStyle}]
 table {%
1 5.067287921905518
2 5.071765899658203
3 4.985106945037842
4 5.094770431518555
5 5.111896514892578
6 5.060968399047852
7 5.028281211853027
8 4.928787708282471
9 4.755165100097656
10 4.611979961395264
11 4.675789833068848
12 4.690364837646484
};
\addlegendentry{iBOT~\cite{zhou2022image}}

 \addplot[{MoCoStyle}]
 table {%
1 4.4867634773254395
2 4.451283931732178
3 4.83425760269165
4 4.610452651977539
5 4.852628231048584
6 4.983875751495361
7 4.976497650146484
8 4.9116950035095215
9 4.879249095916748
10 4.72769021987915
11 4.8500237464904785
12 5.0912322998046875
};
\addlegendentry{MoCo-v3~\cite{chen2021empirical}}

\addplot[{MAEStyle}]
 table {%
1 5.130588531494141
2 5.006633281707764
3 4.973138332366943
4 5.115799903869629
5 5.107693672180176
6 5.029148578643799
7 5.141277313232422
8 5.184921741485596
9 5.1066670417785645
10 5.140458106994629
11 5.05302095413208
12 4.7051873207092285
};
\addlegendentry{MAE~\cite{he2021masked}}

 \addplot[{FinetunedMAEStyle}]
 table {%
1 5.0921430587768555
2 4.984069347381592
3 4.820669174194336
4 4.961766242980957
5 4.982217311859131
6 5.006401538848877
7 5.0303473472595215
8 4.7251739501953125
9 4.540709018707275
10 4.505160331726074
11 4.464832782745361
12 4.398733139038086
};
\addlegendentry{MAE (FT)~\cite{he2021masked}}

\end{axis}
\end{tikzpicture}
    \end{minipage}
    \\
    \begin{minipage}{0.48\linewidth}
        \centering
            \begin{tikzpicture}
        \begin{axis}[
            sec4style,
            ylabel={{\shortstack{Entropy of attention between\\the \cls{}and patch tokens}}},
	xlabel={ViT-B block}
	]
 \addplot[{DINOStyle}]
 table {%
1 5.142787933349609
2 4.672569274902344
3 4.999399185180664
4 5.003344535827637
5 4.872176170349121
6 4.846645832061768
7 4.68649959564209
8 4.586240768432617
9 4.39156436920166
10 4.322977542877197
11 4.356078147888184
12 4.561920642852783
};
\addlegendentry{DINO~\cite{caron2021emerging}}

 \addplot[{iBOTStyle}]
 table {%
1 5.136229515075684
2 4.9121623039245605
3 4.688700199127197
4 5.02535343170166
5 4.955214500427246
6 4.915258884429932
7 4.813562393188477
8 4.644126892089844
9 4.465449333190918
10 4.3610382080078125
11 4.5107502937316895
12 4.679174423217773
};
\addlegendentry{iBOT~\cite{zhou2022image}}

 \addplot[{MoCoStyle}]
 table {%
1 4.251352787017822
2 4.919338703155518
3 4.876862525939941
4 5.0949177742004395
5 5.020232200622559
6 4.991518974304199
7 4.926223278045654
8 4.838733673095703
9 4.690496444702148
10 4.553460597991943
11 4.596993446350098
12 4.623865127563477
};
\addlegendentry{MoCo-v3~\cite{chen2021empirical}}

 \addplot[{MAEStyle}]
 table {%
1 5.036749362945557
2 4.80377197265625
3 4.993056774139404
4 5.023566722869873
5 5.124216556549072
6 5.077910423278809
7 5.099809646606445
8 5.040615081787109
9 5.039320468902588
10 5.027137279510498
11 5.051735877990723
12 5.025387287139893
};
\addlegendentry{MAE~\cite{he2021masked}}

 \addplot[{FinetunedMAEStyle}]
 table {%
1 5.05247163772583
2 4.924233436584473
3 4.977036476135254
4 4.975040435791016
5 5.07895565032959
6 5.055420875549316
7 4.958649635314941
8 4.9717698097229
9 4.844841003417969
10 4.596076488494873
11 4.419389247894287
12 4.2843475341796875
};
\addlegendentry{MAE (FT)~\cite{he2021masked}}

\end{axis}
\end{tikzpicture}
    \end{minipage}
    \hfill
    \begin{minipage}{0.48\linewidth}
        \centering
            \begin{tikzpicture}
        \begin{axis}[
            sec4style,
            ylabel={{\shortstack{Entropy of attention between\\the \cls{}and patch tokens}}},
	xlabel={ViT-L block}
	]
 \addplot[{iBOTStyle}]
 table {%
1 5.067160606384277
2 4.656296730041504
3 4.861199855804443
4 4.878689765930176
5 4.908867359161377
6 4.893306255340576
7 4.845858573913574
8 4.907848834991455
9 4.766055583953857
10 4.722753047943115
11 4.6959147453308105
12 4.595938682556152
13 4.473447322845459
14 4.43056058883667
15 4.296781539916992
16 4.253748893737793
17 4.189476013183594
18 4.128510475158691
19 4.172135829925537
20 4.364649772644043
21 4.467584609985352
22 4.456131935119629
23 4.504015922546387
24 4.51623010635376
};
\addlegendentry{iBOT~\cite{zhou2022image}}

 \addplot[{MAEStyle}]
 table {%
1 5.02154016494751
2 4.825206756591797
3 4.847005844116211
4 4.975472927093506
5 5.068882465362549
6 5.069686412811279
7 4.98040771484375
8 5.052735805511475
9 5.058859825134277
10 4.9680633544921875
11 5.078675270080566
12 5.106662273406982
13 5.068673133850098
14 5.043344497680664
15 5.06113862991333
16 5.049450397491455
17 5.050853252410889
18 5.078267574310303
19 5.075350761413574
20 5.080392837524414
21 5.057682037353516
22 5.052903175354004
23 5.058250427246094
24 5.022083282470703
};
\addlegendentry{MAE~\cite{he2021masked}}

 \addplot[{FinetunedMAEStyle}]
 table {%
1 5.006109237670898
2 4.8454670906066895
3 4.79143762588501
4 4.913954734802246
5 5.02326774597168
6 5.006096363067627
7 4.923790454864502
8 4.962786674499512
9 5.00257682800293
10 4.8624067306518555
11 4.959377288818359
12 4.974221706390381
13 4.948400020599365
14 4.834353923797607
15 4.821866989135742
16 4.698546886444092
17 4.640446186065674
18 4.513915538787842
19 4.392924785614014
20 4.191298007965088
21 4.160512924194336
22 4.093625545501709
23 4.067463397979736
24 4.179765224456787
};
\addlegendentry{MAE (FT)~\cite{he2021masked}}

\end{axis}
\end{tikzpicture}
    \end{minipage}
    \caption{
    Extended version of \Cref{fig:cls_pos_entropy}.
    Entropy of attention between the \cls{} and patch
tokens. In MAE, its value reaches almost the maximal possible
level, In other models, it decreases in the deeper model blocks,
indicating that the \cls{} token attends to different patches in a
more selective manner. Fine-tuning of MAE decreases this entropy.
indicating that selective attention to patch tokens is crucial for good
perception.
    }
    \label{fig:cls_pos_entropy_all_vits}
\end{figure*}

\begin{figure*}[t]
    \centering
    \begin{minipage}{0.48\linewidth}
        \begin{tikzpicture}[
    node distance=1.25cm and 0.75cm,
]

    \node[draw, gray] (zin0) {$\mathbf{v}_0$};
    \node[draw, right=1 cm of zin0] (zin1) {$\mathbf{v}_1$};
    \node[right=0.1cm of zin1] (zindots) {$\dots$};
    \node[draw, right=0.1 cm of zindots] (zini) {$\mathbf{v}_i$};
    \node[right=0.1 cm of zini] (zindots2) {$\dots$};

    \node[draw, right=0.1 cm of zindots2] (zinN) {$\mathbf{v}_N$};

    \draw[decorate,decoration={brace,amplitude=6pt}, gray] 
        ([yshift=5pt]zin0.north west) -- ([yshift=5pt]zin0.north east) 
        node[midway, above=6pt] {\cls{} token};

    \draw[decorate,decoration={brace,amplitude=6pt}] 
        ([yshift=5pt]zin1.north west) -- ([yshift=5pt]zinN.north east) 
        node[midway, above=6pt] {Patch tokens};

    \node[left=of zin0, align=center] (int) {Value\\tokens};
    \node[below=of int, align=center] (outt) {Output\\tokens};
    \draw[-stealth, thick] (int) -- (outt) node[midway, fill=white, inner sep=1pt,align=center] {\scriptsize Attention};

    \node[draw, below=1.7 cm of zini] (zouti) {$\mathbf{o}_{l}^i$};

    \draw[-stealth, thick, gray] (zin0.south) -- (zouti) node[midway, fill=white, inner sep=0pt] {$\mathbf{a}_{i,0}$};
    \draw[-stealth, thick] (zin1) -- (zouti) node[midway, fill=white, inner sep=2pt] (ai1) {$\mathbf{a}_{i,1}$};
    \draw[-stealth, thick] (zini) -- (zouti) node[midway, fill=white, inner sep=3pt, draw=red, very thick, dashed, rounded corners] (aii) {$\mathbf{a}_{i,i}$};

    \draw[-stealth, thick] (zinN) -- (zouti) node[midway, fill=white, inner sep=2pt] (aiN) {$\mathbf{a}_{i,N}$};

    \node[draw=blue, dashed,fit=(ai1) (aiN), very thick, inner sep=3pt, rounded corners] (a_container) {};

    \node[below left= 0.4 cm and 0 cm of a_container, inner sep=-1.5pt] (ratio) {$\dfrac{\textcolor{red}{\mathlarger{\mathlarger{\bullet}}}}{\Sigma \textcolor{blue}{\mathlarger{\mathlarger{\bullet}}}}$};
    \draw[-stealth, very thick, dashed, red] (aii) -- (ratio.north east);
    \draw[-stealth, very thick, dashed, blue] (a_container) -- (ratio.south east);

    \node[below =1.9 cm of zindots2] {$\dots$};

\end{tikzpicture}
    \end{minipage}
    \hfill
    \begin{minipage}{0.48\linewidth}
        \centering
    \begin{tikzpicture}
        \begin{axis}[
            sec4style,
            ylabel={{\shortstack{Normalized attention
of patch tokens to themselves}}},
	xlabel={ViT-S block}
	]
 \addplot[{DINOStyle}]
 table {%
1 0.06217394024133682
2 0.1361553966999054
3 0.06116822734475136
4 0.04480546712875366
5 0.051384586840867996
6 0.03232467547059059
7 0.028756100684404373
8 0.02395019121468067
9 0.0136968232691288
10 0.009706471115350723
11 0.00801673624664545
12 0.0076947445049881935
};
\addlegendentry{DINO~\cite{caron2021emerging}}

 \addplot[{iBOTStyle}]
 table {%
1 0.10080636292695999
2 0.10572224855422974
3 0.06164490059018135
4 0.057509955018758774
5 0.0687771886587143
6 0.046239569783210754
7 0.03621291741728783
8 0.03131624683737755
9 0.022473884746432304
10 0.011617018841207027
11 0.02004241570830345
12 0.03146136552095413
};
\addlegendentry{iBOT~\cite{zhou2022image}}

 \addplot[{MoCoStyle}]
 table {%
1 0.013782881200313568
2 0.04465027526021004
3 0.09547702968120575
4 0.06620177626609802
5 0.1108965203166008
6 0.11679747700691223
7 0.06594535708427429
8 0.05709525942802429
9 0.023073075339198112
10 0.01189368311315775
11 0.009083683602511883
12 0.006792111322283745
};
\addlegendentry{MoCo-v3~\cite{chen2021empirical}}

\addplot[{MAEStyle}]
 table {%
1 0.10564249753952026
2 0.06675615161657333
3 0.027471201494336128
4 0.03598170727491379
5 0.15919733047485352
6 0.05007298290729523
7 0.14304664731025696
8 0.05095616355538368
9 0.04929615929722786
10 0.032668665051460266
11 0.035886507481336594
12 0.03599434718489647
};
\addlegendentry{MAE~\cite{he2021masked}}

 \addplot[{FinetunedMAEStyle}]
 table {%
1 0.11491706967353821
2 0.09745321422815323
3 0.047169797122478485
4 0.04538021981716156
5 0.1874724179506302
6 0.04971252381801605
7 0.18265201151371002
8 0.029400493949651718
9 0.018870333209633827
10 0.01794838346540928
11 0.012345273047685623
12 0.014299549162387848
};
\addlegendentry{MAE (FT)~\cite{he2021masked}}

\end{axis}
\end{tikzpicture}
    \end{minipage}
    \begin{minipage}{0.48\linewidth}
        \centering
    \begin{tikzpicture}
        \begin{axis}[
            sec4style,
            ylabel={{\shortstack{Relative attention between\\the same patch tokens}}},
	xlabel={ViT-B block}
	]
 \addplot[{DINOStyle}]
 table {%
1 0.10427000373601913
2 0.058244701474905014
3 0.04074423387646675
4 0.03401390463113785
5 0.031960275024175644
6 0.021716957911849022
7 0.019277183338999748
8 0.018366316333413124
9 0.011647745966911316
10 0.011410688050091267
11 0.009314093738794327
12 0.011983493342995644
};
\addlegendentry{DINO~\cite{caron2021emerging}}

 \addplot[{iBOTStyle}]
 table {%
1 0.08299794793128967
2 0.08324775099754333
3 0.06118907406926155
4 0.03964445739984512
5 0.04465824365615845
6 0.029700549319386482
7 0.0255900751799345
8 0.023498745635151863
9 0.017476389184594154
10 0.012934492900967598
11 0.02271910198032856
12 0.026191875338554382
};
\addlegendentry{iBOT~\cite{zhou2022image}}

 \addplot[{MoCoStyle}]
 table {%
1 0.012961355037987232
2 0.07047972828149796
3 0.04581468924880028
4 0.05335165560245514
5 0.053382404148578644
6 0.03896085172891617
7 0.03106638975441456
8 0.03525640442967415
9 0.01549577433615923
10 0.014685172587633133
11 0.008616182953119278
12 0.007575439289212227
};
\addlegendentry{MoCo-v3~\cite{chen2021empirical}}

\addplot[{MAEStyle}]
 table {%
1 0.030649458989501
2 0.07072880864143372
3 0.04010970890522003
4 0.033401232212781906
5 0.09110450744628906
6 0.049564212560653687
7 0.10953568667173386
8 0.07168903201818466
9 0.030591905117034912
10 0.03453614562749863
11 0.025126393884420395
12 0.025260325521230698
};
\addlegendentry{MAE~\cite{he2021masked}}

 \addplot[{FinetunedMAEStyle}]
 table {%
1 0.038292743265628815
2 0.08017653971910477
3 0.04946440830826759
4 0.045961491763591766
5 0.08078787475824356
6 0.03815276175737381
7 0.036439187824726105
8 0.02027437463402748
9 0.01154826395213604
10 0.0073655713349580765
11 0.0151986638084054
12 0.016380617395043373
};
\addlegendentry{MAE (FT)~\cite{he2021masked}}

\end{axis}
\end{tikzpicture}
    \end{minipage}
    \hfill
    \begin{minipage}{0.48\linewidth}
        \centering
            \begin{tikzpicture}
        \begin{axis}[
            sec4style,
            ylabel={{\shortstack{Relative attention between\\the same patch tokens}}},
	xlabel={ViT-L block}
	]
 \addplot[{iBOTStyle}]
 table {%
1 0.07621852308511734
2 0.022609282284975052
3 0.033166103065013885
4 0.03206636384129524
5 0.030449599027633667
6 0.03609975427389145
7 0.02632358856499195
8 0.026424404233694077
9 0.023278124630451202
10 0.024250397458672523
11 0.019645243883132935
12 0.022312838584184647
13 0.013504943810403347
14 0.014377154409885406
15 0.009606216102838516
16 0.009025686420500278
17 0.008913813158869743
18 0.009508255869150162
19 0.010083289816975594
20 0.013067317195236683
21 0.020447829738259315
22 0.0283469557762146
23 0.02926577441394329
24 0.033707715570926666
};
\addlegendentry{iBOT~\cite{zhou2022image}}

 \addplot[{MAEStyle}]
 table {%
1 0.017023319378495216
2 0.06975562125444412
3 0.032907430082559586
4 0.02629750780761242
5 0.0355071984231472
6 0.032169029116630554
7 0.031115302816033363
8 0.08934631943702698
9 0.12116339802742004
10 0.0337241105735302
11 0.09460954368114471
12 0.11782553046941757
13 0.05027543380856514
14 0.08436954766511917
15 0.03537401929497719
16 0.0320449061691761
17 0.03687620535492897
18 0.02539176493883133
19 0.04386788606643677
20 0.03994158282876015
21 0.04059361293911934
22 0.03273608162999153
23 0.03482447564601898
24 0.040369126945734024
};
\addlegendentry{MAE~\cite{he2021masked}}

 \addplot[{FinetunedMAEStyle}]
 table {%
1 0.01713782548904419
2 0.08377217501401901
3 0.03974304348230362
4 0.02884184755384922
5 0.04030058532953262
6 0.034212831407785416
7 0.03142622113227844
8 0.09083137661218643
9 0.12853682041168213
10 0.03373047709465027
11 0.09213121235370636
12 0.10042671114206314
13 0.030418740585446358
14 0.06953797489404678
15 0.016298100352287292
16 0.011688734404742718
17 0.01047238614410162
18 0.00598008232191205
19 0.009861310012638569
20 0.012983549386262894
21 0.01133433822542429
22 0.011265481822192669
23 0.014770963229238987
24 0.009102551266551018
};
\addlegendentry{MAE (FT)~\cite{he2021masked}}

\end{axis}
\end{tikzpicture}
    \end{minipage}

    \caption{
    Extended version of \Cref{fig:pos_self_attn_adj}.
     Attention of the patch tokens to themselves, relative to the total attention assigned to all patch tokens. In the latter MAE blocks, patch tokens seem to assign the largest amount of relative attention to themselves, compared to the tokens of JEA.}
    \label{fig:pos_self_attn_adj_all_vits}
\end{figure*}

\begin{figure*}[t]
    \centering
    \begin{minipage}{0.48\linewidth}
        \centering
        \begin{tikzpicture}[
    node distance=1.25cm and 0.75cm,
]

    \node[draw, gray] (zin0) {$\mathbf{v}_0$};
    \node[draw, right=1 cm of zin0] (zin1) {$\mathbf{v}_1$};
    \node[right=0.1cm of zin1] (zindots) {$\dots$};
    \node[draw, right=0.1 cm of zindots] (zini) {$\mathbf{v}_i$};
    \node[right=0.1 cm of zini] (zindots2) {$\dots$};

    \node[draw, right=0.1 cm of zindots2] (zinN) {$\mathbf{v}_N$};

    \draw[decorate,decoration={brace,amplitude=6pt}, gray] 
        ([yshift=5pt]zin0.north west) -- ([yshift=5pt]zin0.north east) 
        node[midway, above=6pt] {\cls{} token};

    \draw[decorate,decoration={brace,amplitude=6pt}] 
        ([yshift=5pt]zin1.north west) -- ([yshift=5pt]zinN.north east) 
        node[midway, above=6pt] {Patch tokens};

    \node[left=of zin0, align=center] (int) {Value\\tokens};
    \node[below=of int, align=center] (outt) {Output\\tokens};
    \draw[-stealth, thick] (int) -- (outt) node[midway, fill=white, inner sep=1pt,align=center] {\scriptsize Attention};

    \node[draw, below=1.7 cm of zini] (zouti) {$\mathbf{o}_i$};

    \draw[-stealth, thick, gray] (zin0.south) -- (zouti) node[midway, fill=white, inner sep=0pt] {$\mathbf{a}_{i,0}$};
    \draw[-stealth, thick] (zin1) -- (zouti) node[midway, fill=white, inner sep=2pt] (ai1) {$\mathbf{a}_{i,1}$};
    \draw[-stealth, thick] (zini) -- (zouti) node[midway, fill=white, inner sep=2pt] (aii) {$\mathbf{a}_{i,i}$};

    \draw[-stealth, thick] (zinN) -- (zouti) node[midway, fill=white, inner sep=2pt] (aiN) {$\mathbf{a}_{i,N}$};

    \node[draw=red, dashed,fit=(ai1) (aiN), very thick, inner sep=2pt, rounded corners] (a_container) {};

    \node[below right= 0.25 cm and -0.4 cm of a_container, red] (label) {Entropy};
    \draw[-stealth, very thick, dashed, red] (a_container) -- (label);

    \node[below=1.85cm of zindots] {$\dots$};

\end{tikzpicture}
    \end{minipage}
    \hfill
    \begin{minipage}{0.48\linewidth}
        \centering
            \begin{tikzpicture}
        \begin{axis}[
            sec4style,
            ylabel={{\shortstack{Entropy of normalized attention
between patch tokens}}},
	xlabel={ViT-S block}
	]
 \addplot[{DINOStyle}]
 table {%
1 4.721439361572266
2 4.371654510498047
3 3.3631999492645264
4 3.9588916301727295
5 3.5877928733825684
6 4.219051361083984
7 4.181548595428467
8 4.146601676940918
9 4.515894412994385
10 4.819899559020996
11 4.984620571136475
12 5.119161605834961
};
\addlegendentry{DINO~\cite{caron2021emerging}}

 \addplot[{iBOTStyle}]
 table {%
1 4.442992210388184
2 4.416973114013672
3 2.606908082962036
4 3.4099862575531006
5 3.6504149436950684
6 4.009920120239258
7 4.075101852416992
8 4.156713008880615
9 4.305519104003906
10 4.695773601531982
11 4.860165119171143
12 4.838006019592285
};
\addlegendentry{iBOT~\cite{zhou2022image}}

 \addplot[{MoCoStyle}]
 table {%
1 4.3098835945129395
2 4.4070143699646
3 4.165054798126221
4 3.364975929260254
5 3.372143268585205
6 3.798346757888794
7 4.263214588165283
8 4.120254993438721
9 4.650585651397705
10 4.742236614227295
11 4.925751209259033
12 5.243824481964111
};
\addlegendentry{MoCo-v3~\cite{chen2021empirical}}

 \addplot[{MAEStyle}]
 table {%
1 4.306061744689941
2 4.5760602951049805
3 4.488229751586914
4 4.022392272949219
5 1.738390564918518
6 2.7743256092071533
7 3.0250229835510254
8 3.814295530319214
9 3.8428890705108643
10 4.002144813537598
11 3.9560744762420654
12 4.030394077301025
};
\addlegendentry{MAE~\cite{he2021masked}}

 \addplot[{FinetunedMAEStyle}]
 table {%
1 4.194010257720947
2 4.348647117614746
3 4.37178897857666
4 3.8198978900909424
5 1.8478589057922363
6 2.963932991027832
7 3.055875301361084
8 3.8676488399505615
9 4.10698127746582
10 4.154465675354004
11 4.369101047515869
12 4.306178569793701
};
\addlegendentry{MAE (FT)~\cite{he2021masked}}

\end{axis}
\end{tikzpicture}
    \end{minipage}
    \begin{minipage}{0.48\linewidth}
        \centering
            \begin{tikzpicture}
        \begin{axis}[
            sec4style,
            ylabel={{\shortstack{Entropy of attention\\between patch tokens}}},
	xlabel={ViT-B block}
	]
 \addplot[{DINOStyle}]
 table {%
1 4.3837690353393555
2 3.6690053939819336
3 3.6203320026397705
4 3.6401712894439697
5 3.7775065898895264
6 4.022028923034668
7 4.013164043426514
8 4.112133979797363
9 4.503214359283447
10 4.712254524230957
11 4.827049732208252
12 4.955743312835693
};
\addlegendentry{DINO~\cite{caron2021emerging}}

 \addplot[{iBOTStyle}]
 table {%
1 4.466940879821777
2 3.658950090408325
3 3.2001729011535645
4 3.753647565841675
5 3.8279638290405273
6 4.036083698272705
7 4.13369083404541
8 4.075435638427734
9 4.273280143737793
10 4.579384803771973
11 4.731393814086914
12 4.682201862335205
};
\addlegendentry{iBOT~\cite{zhou2022image}}

 \addplot[{MoCoStyle}]
 table {%
1 4.0157294273376465
2 4.436943054199219
3 3.4602842330932617
4 3.637147903442383
5 3.8815367221832275
6 4.102165222167969
7 4.23651123046875
8 4.000644207000732
9 4.430052757263184
10 4.479116916656494
11 4.757903575897217
12 5.199434757232666
};
\addlegendentry{MoCo-v3~\cite{chen2021empirical}}

 \addplot[{MAEStyle}]
 table {%
1 4.718160629272461
2 4.302433490753174
3 4.403627395629883
4 4.527017593383789
5 2.737722873687744
6 3.0187857151031494
7 2.9202182292938232
8 3.9761245250701904
9 4.13281774520874
10 4.054203510284424
11 4.30564022064209
12 4.246712684631348
};
\addlegendentry{MAE~\cite{he2021masked}}

 \addplot[{FinetunedMAEStyle}]
 table {%
1 4.532347679138184
2 4.233613967895508
3 4.2133469581604
4 4.285141468048096
5 2.9295084476470947
6 3.644650459289551
7 3.7847113609313965
8 4.129693031311035
9 4.158210754394531
10 4.082627773284912
11 3.9730658531188965
12 3.981553316116333
};
\addlegendentry{MAE (FT)~\cite{he2021masked}}

\end{axis}
\end{tikzpicture}
    \end{minipage}
    \hfill
    \begin{minipage}{0.48\linewidth}
        \centering
            \begin{tikzpicture}
        \begin{axis}[
            sec4style,
            ylabel={{\shortstack{Entropy of attention\\between patch tokens}}},
	xlabel={ViT-L block}
	]
 \addplot[{iBOTStyle}]
 table {%
1 4.367062568664551
2 4.348813056945801
3 3.449146270751953
4 3.130492925643921
5 3.9601309299468994
6 3.71994948387146
7 3.8310015201568604
8 3.925179958343506
9 3.8333992958068848
10 3.877959966659546
11 3.9731388092041016
12 3.867311954498291
13 4.1474223136901855
14 4.230665683746338
15 4.465093612670898
16 4.522359371185303
17 4.615249156951904
18 4.661001682281494
19 4.70122766494751
20 4.676745891571045
21 4.613316059112549
22 4.537418842315674
23 4.458315849304199
24 4.4570488929748535
};
\addlegendentry{iBOT~\cite{zhou2022image}}

 \addplot[{MAEStyle}]
 table {%
1 4.836921691894531
2 3.9356846809387207
3 4.409603595733643
4 4.735604286193848
5 4.446406841278076
6 3.460235357284546
7 3.596498727798462
8 3.448171615600586
9 3.294121026992798
10 3.77762508392334
11 3.2896244525909424
12 3.1579647064208984
13 3.369508981704712
14 3.890223503112793
15 3.968224287033081
16 4.070300102233887
17 3.9230620861053467
18 4.359838962554932
19 3.9235332012176514
20 4.003920555114746
21 4.034440040588379
22 3.9728705883026123
23 4.321606159210205
24 4.12828254699707
};
\addlegendentry{MAE~\cite{he2021masked}}

 \addplot[{FinetunedMAEStyle}]
 table {%
1 4.809374809265137
2 3.863128185272217
3 4.335094928741455
4 4.6601104736328125
5 4.329829216003418
6 3.3689353466033936
7 3.5009098052978516
8 3.3230652809143066
9 3.1231377124786377
10 3.6012649536132812
11 3.2889978885650635
12 3.2139382362365723
13 3.4390366077423096
14 3.674266815185547
15 3.789095878601074
16 3.7883615493774414
17 3.7361178398132324
18 3.5948305130004883
19 3.567811965942383
20 3.456134080886841
21 3.464481830596924
22 3.065232753753662
23 3.6495015621185303
24 4.470020771026611
};
\addlegendentry{MAE (FT)~\cite{he2021masked}}

\end{axis}
\end{tikzpicture}
    \end{minipage}
    \caption{
    Extended version of \Cref{fig:pos_pos_entropy}.
    Entropy of attention of patch tokens to patch tokens. 
    In MAE, the patch tokens attend to other patches with lower entropy than in JEA, suggesting that they form a representation of their local image fragments. 
    }
    \label{fig:pos_pos_entropy_all_vits}
\end{figure*}
    \pgfplotsset{ylabel/.style={\pgfkeysvalueof{/pgfplots/ylabel}}}


\end{document}